\documentclass[lettersize,journal]{IEEEtran}
\usepackage{adjustbox}
\usepackage{graphicx}
\usepackage{amsmath}
\usepackage{amssymb}
\usepackage{booktabs}
\usepackage{color}
\usepackage{epsfig}
\usepackage{graphicx}
\usepackage{amsmath}
\usepackage{verbatim}
\usepackage{amssymb}
\usepackage{bm}
\usepackage{multirow}
\usepackage{booktabs}
\usepackage{tabularx}
\usepackage{url}
\usepackage{graphicx}
\usepackage{mathrsfs}
\usepackage{amsmath}
\usepackage{multirow}
\usepackage{rotfloat}
\usepackage{hhline}
\usepackage{graphicx}
\usepackage[table,xcdraw]{xcolor}
\usepackage[normalem]{ulem}
\useunder{\uline}{\ul}{}
\usepackage{tablefootnote}
\usepackage{bbding}
\usepackage{pifont}
\usepackage{wasysym}
\usepackage[table,xcdraw]{xcolor}
\usepackage[colorlinks,linkcolor=red,anchorcolor=green,citecolor=blue]{hyperref}

\usepackage{amsmath,amsfonts}
\usepackage{algorithmic}
\usepackage{algorithm}
\usepackage{array}
\usepackage[caption=false,font=normalsize,labelfont=sf,textfont=sf]{subfig}
\usepackage{textcomp}
\usepackage{stfloats}
\usepackage{url}
\usepackage{verbatim}
\usepackage{graphicx}
\usepackage{cite}
\begin{document}
	
	\title{Infrared and Visible Image Fusion: From Data Compatibility to Task Adaption}
	
	\author{	
		Jinyuan~Liu,~\IEEEmembership{Member,~IEEE,} Guanyao Wu, Zhu Liu, Di Wang, Zhiying Jiang, Long Ma, ~\IEEEmembership{Member,~IEEE,} Wei Zhong, Xin~Fan,~\IEEEmembership{Senior Member,~IEEE,}  Risheng~Liu,~\IEEEmembership{Member,~IEEE,}
		
		\IEEEcompsocitemizethanks{\IEEEcompsocthanksitem 
		This work is supported in part by the National Natural Science Foundation of China (Nos. U22B2052, 62450072, 12326605, 62027826, 62302078, 62372080); in part by National Key Research and Development Program of China (No. 2022YFA1004101); in part by China Postdoctoral Science Foundation (No.
		2023M730741); and in part by the Fundamental Research Funds for the Central Universities (Nos. DUT24BS013, DUT24LAB125).
		
		Jinyuan Liu, Guanyao Wu, Zhu Liu, Di Wang, Zhijing Jiang, Long Ma, Wei Zhong, Xin Fan and Risheng Liu are with the School of Software Technology, Dalian University of Technology, Dalian, 116024, China.
		(\textit{Corresponding author: Risheng Liu, e-mail: rsliu@dlut.edu.cn}	).	
		}
	}
	
	\markboth{Journal of \LaTeX\ Class Files,~Vol.~14, No.~8, August~2021}
	{Shell \MakeLowercase{\textit{et al.}}: A Sample Article Using IEEEtran.cls for IEEE Journals}

	\maketitle
	
	\begin{abstract}
		Infrared-visible image fusion (IVIF) is a fundamental and critical task in the field of computer vision. Its aim is to integrate the unique characteristics of both infrared and visible spectra into a holistic representation. Since 2018,  growing amount and diversity IVIF approaches step into a deep-learning era, encompassing introduced a broad spectrum of networks or loss functions for improving visual enhancement. As research deepens and practical demands grow, several intricate issues like data compatibility, perception accuracy, and efficiency cannot be ignored. Regrettably, there is a lack of recent surveys that comprehensively introduce and organize this expanding domain of knowledge. Given the current rapid development, this paper aims to fill the existing gap by providing a comprehensive survey  that covers a wide array of aspects. Initially, we introduce a multi-dimensional framework to elucidate the prevalent learning-based IVIF methodologies, spanning topics from basic visual enhancement strategies to data compatibility, task adaptability, and further extensions. Subsequently, we delve into a profound analysis of these new approaches,  offering a detailed lookup table to clarify their core ideas.  Last but not the least, We also summarize performance comparisons quantitatively and qualitatively, covering registration, fusion and follow-up high-level tasks. Beyond delving into the technical nuances of these learning-based fusion approaches, we also explore potential future directions and open issues that warrant further exploration by the community. For additional information and a detailed data compilation, please refer to our GitHub repository: \url{https://github.com/RollingPlain/IVIF_ZOO}. 
	\end{abstract}
	
	
	\section{Introduction}
	
\begin{figure*}[!htb]
		\centering
		\setlength{\tabcolsep}{1pt} 
		
		\includegraphics[width=0.98\textwidth]{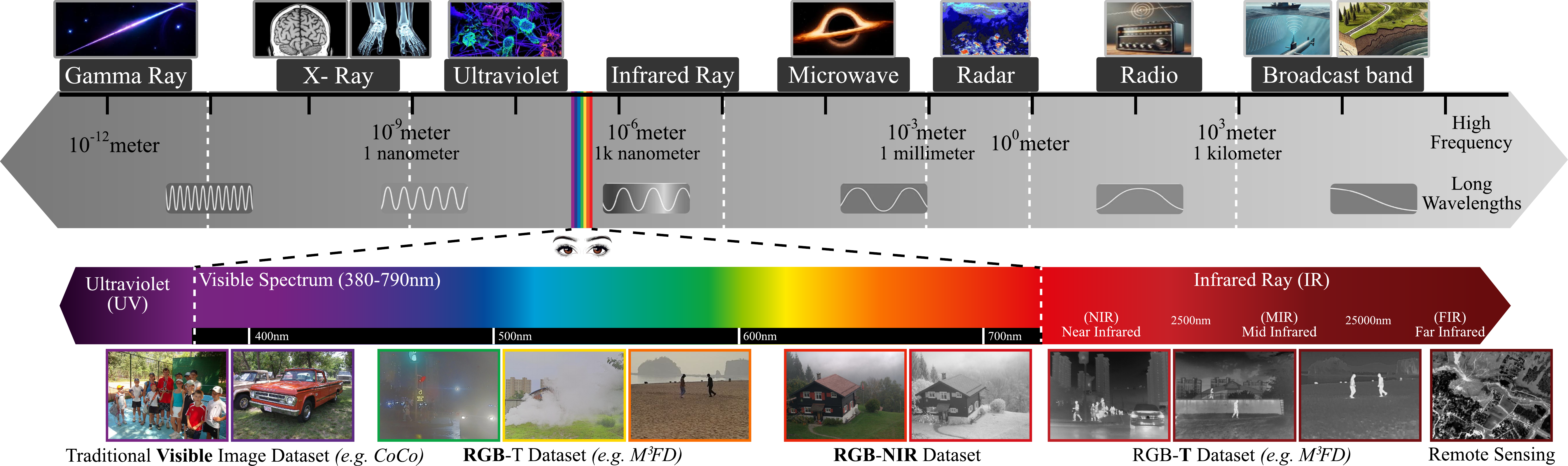}
		\vspace{-0.2cm}
		\caption{A detailed spectrogram depicting almost all wavelength and frequency ranges, particularly expanding the range of the human visual system and annotating corresponding computer vision and image fusion datasets~\cite{lin2014microsoft}.}
		\label{fig: Spectrogram}
		\vspace{-0.5cm}
	\end{figure*}

	\IEEEPARstart{I} n spectral terms, a spectrum can be succinctly defined as the representation of light across varying frequencies or wavelengths. Broadly speaking, the spectrum encompasses the entire electromagnetic range, spanning from radio waves to gamma rays. However, in our daily lives, we are most familiar with the visible spectrum, which includes colors perceptible to the human eye, such as red, orange, yellow, green, blue, indigo, and violet.
	Yet, the visible spectrum represents only a minuscule portion of the entire spectrum. Beyond the realm of visible light, there exist numerous other spectral types, including ultraviolet, infrared, microwaves, and X-rays, each with its distinct characteristics and applications~\cite{marpaung2019integrated, simon2005microwave}. Figure.~\ref{fig: Spectrogram} displays images corresponding to different wavelengths, ranging from $10^{-12}$ meter~(gamma ray) to $10^{3}$ meter~(broadcast band).
	
	With rapid advancement of artificial intelligence, there is a growing perceptual demand for intelligent systems to operate effectively in real-world, complex, and even extreme scenarios. However, the inherent limitations of singe spectral type often prevent a holistic, trustworthy, and precise depiction of these scenarios. Consequently, multi-spectral image fusion
	technologies have emerged, aiming to synthesize and refine the diverse image data from different sensors capturing the same environment~\cite{simone2002image, yokoya2017hyperspectral}.
	Among these, infrared and visible images act as the primary data sources for next-generation intelligent systems, playing an irreplaceable role in achieving high-reliability perception tasks.

	The infrared spectrum lies just outside the red end of the visible spectrum, possessing wavelengths longer than red light. One of the salient features of infrared radiation is its ability to be absorbed and re-emitted by objects, making it an invaluable indicator of temperature or heat. Consequently, infrared sensors are frequently employed for night vision, thermal imaging, and certain specialized medical applications~\cite{li2023simulation}. While both spectral bands have constraints in adverse conditions, multi-spectral imagery offers significant complementarity in terms of environmental adaptability (e.g., smoke, obstructions, low light) and distinct visual features (e.g., resolution, contrast, texture detail). 
	{To build on the complementarity of infrared and visible spectra, it is essential to integrate the two to leverage their combined strengths. A straightforward way is to feed both infrared and visible images directly into neural networks, allowing for decision-level fusion to accomplish various tasks. However, this method overlooks the advantages of generating a fused image, which can enhance information representation, reduce noise, and simultaneously satisfy observational requirements while better supporting wide practical applications, such as remote sensing, military surveillance, and autonomous driving, to name a few~\cite{liu2015general}. }

	Since 2018, learning-based IVIF approaches have undergone substantial development due to the robust non-linear fitting capabilities provided by deep learning techniques. These learning-based solutions, compared to their conventional counterparts, exhibit superior visual quality, robustness, and computational efficiency, thereby attracting increasing attention. In terms of these approaches, they often achieve the state-of-the-art performance
	on various benchmarks, ranging from the early image fusion approaches for only visual enhancement to recent data compatibility/task adaptability approaches.

	This manuscript provides a comprehensive overview of recent advancements in the fusion of infrared and visible images via deep learning, with a particular emphasis on proposals aimed at practical applications. Although there are existing surveys~\cite{ma2019infrared, zhang2023visible} in the literature concerning infrared and visible image fusion, our work stands out distinctly. Our focus is sharply on IVIF techniques with multi-dimensional insights~(data, fusion and task), in contrast to the majority of previous works that primarily aim to survey either traditional or learning-based IVIF approaches. 
	
	Distinctly, our survey adopts a more general viewpoint, scrutinizing various key factors essential for a deeper understanding of designing deep networks tailored for practical applications. Importantly, we underscore the crucial role of preliminary data compatibility and subsequent tasks, both of which are pivotal when applying IVIF to practical scenarios. Hence, this survey represents the inaugural effort to provide an insightful and systematic analysis of the recent advancements in a multi-dimensional manner. This work is poised to inspire new research endeavors within the community, serving as a catalyst for further exploration and development in the field. The main contributions of this study are fourfold.

\begin{figure}[!htb]
		\centering
		\setlength{\tabcolsep}{1pt} 
		
		\includegraphics[width=0.40\textwidth]{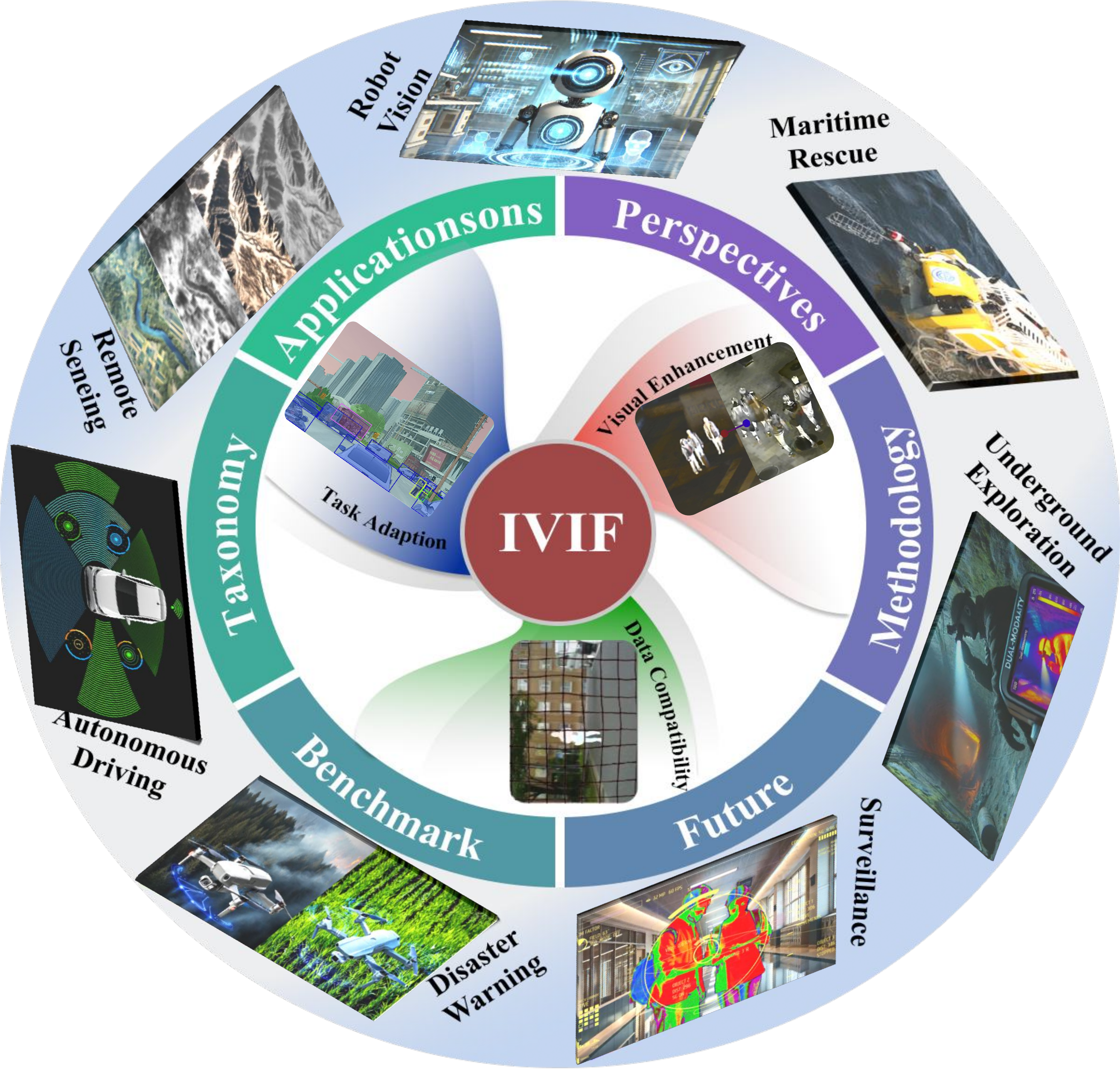}
		\vspace{-0.2cm}
		\caption{A knowledge graph of our survey. We first understand IVIF from three different dimensions, and then we elaborate it on the fusion around seven specific aspects.}
		\label{fig:knowledge}
		\vspace{-0.4cm}
	\end{figure}

	\begin{itemize}
		\item 	
		To the best of our knowledge, this survey paper is unprecedented in its focus on uniformly understanding and organizing learning-based infrared and visible image fusion approaches from a multi-dimensional perspective~(data, fusion and task). We review over 180+ learning-based approaches. A glimpse of the overall structure of the paper is depicted in Figure~\ref{fig:knowledge}.

		\item We engage in a deep discussion regarding each perspective, in terms of recently employed architectures and loss functions. We also summarize a look-up Table~\ref{tab:bigbiao} to discuss the core idea of the representative methods, thereby providing significant convenience for researchers who embark on subsequent studies in this domain.
		
		\item To shed light on application-orient infrared and visible image fusion approaches, we offer a systematic overview  of recent advancements in techniques and datasets in a hierarchical and structured manner. Notably,  we are the first to compare the fusion performance of preliminary registration and subsequent tasks, such as object detection and semantic segmentation.
		
		\item  We address the challenges and open issues, identifying emerging trends and prospective directions to furnish the community with insightful guidance.
	\end{itemize}
	
	\subsection{Scope}
	Image fusion, a foundational image enhancement technique, encompasses various branches, including but not limited to infrared and visible image fusion. Undertaking a detailed review of these interrelated technologies within a single manuscript is impractical. In this work, our primary focus is on learning-based infrared and visible image fusion, with a selective introduction of representative methods from related fields. This paper predominantly centers on significant advancements made over the past six years, with particular attention to works published in top-tier conferences and journals. Besides elucidating the technical details of learning-based infrared and visible image fusion approaches, the survey also outlines taxonomies, popular datasets, potential challenges, and research directions. 
	
	\subsection{Organization}
	The manuscript is organized as follows. Section II provides a concise introduction to the infrared and visible image fusion task with a focus on its applications. In Section III, we introduce a novel taxonomy that categorizes existing approaches into three dimensions. Additionally, this section delves into the fundamental components of network architectures and associated loss functions. This comprehensive overview facilitates beginners in grasping the essentials and aids seasoned researchers in developing a more structured and profound understanding of the domain. Section IV enumerates the widely-accepted benchmarks and evaluation metrics pertinent to the field. Section V showcases a thorough evaluation of our approach, detailing both qualitative and quantitative results across tasks such as registration, fusion, and other subsequent downstream operations. Potential avenues for future research are highlighted in Section VI. The manuscript concludes with a summary in Section VII.
	 
	\vspace{-0.4cm}
	
	\section{Task}
	Given a pair of infrared and visible image captured by different sensors, the goal of IVIF is to generate a signal image that has complementary and comprehensive information than either one.
	Visible images, are generated when sensors utilize light reflecting off various scenes and objects, effectively presenting detailed texture information of the environment. However, these images are easily affected by factors such as ambient lighting, brightness, and adverse weather conditions. In contrast, infrared sensors image through detecting thermal radiation, which accentuates the overall contours of objects, potentially resulting in blurred features, low contrast, and diminished texture information. To capitalize on the strengths of both technologies, a combination of infrared and visible imaging can be used to extract comprehensive information, thereby enhancing scene understanding. This synergy enables practical applications, \emph{e.g.,} intelligent system, in reality to maintain robust visual perception even in environments characterized by dynamic and harsh conditions.
	
	\begin{figure*}[!htb]
		\centering
		\setlength{\tabcolsep}{1pt} 
		
		\includegraphics[width=0.98\textwidth]{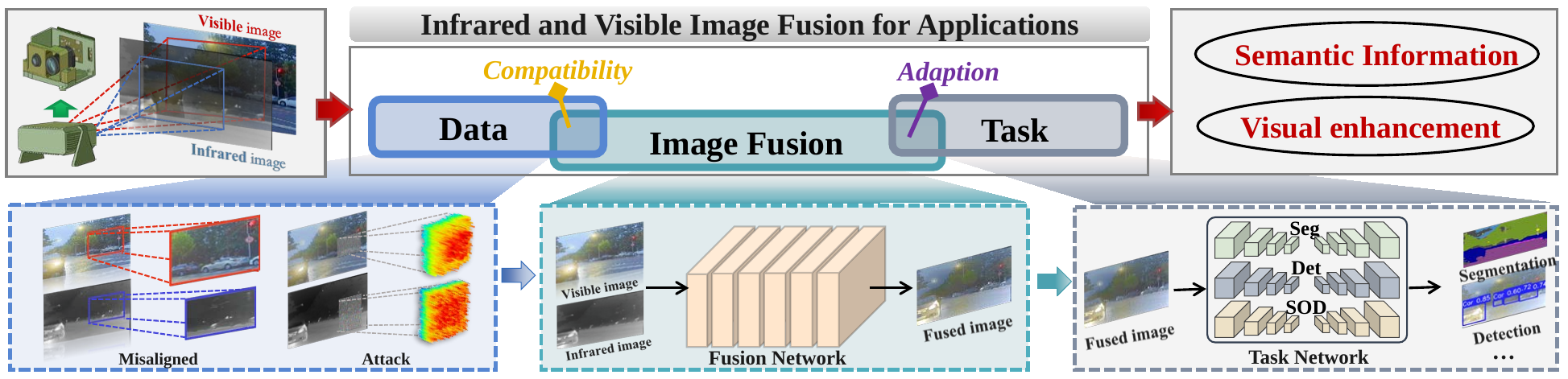}
		\vspace{-0.2cm}
		\caption{{The diagram of infrared and visible image fusion for practical applications. Existing image fusion methods majorly focus on the design of  architectures and training strategies for visual enhancement, few considering the adaptation for downstream visual perception tasks. Additionally, from the data compatibility perspective, pixel misalignment and adversarial attacks of image fusion are two major challenges.  Additionally, integrating comprehensive semantic information for tasks like semantic segmentation, object detection, and salient object detection remains underexplored, posing a critical obstacle in  image fusion.}}
		\label{fig:pipeline}

	\end{figure*}
	
	When applying IVIF to practical applications, two key factors should be careful considered. (i)~Most of existing approaches need pixel-level registered infrared and visible image pairs for fusion. However, due to the significant differences in the viewpoints, pixel distributions, and resolutions of the infrared and visible senors, obtaining accurately registered data is extremely challenging. (ii)~A part of approaches focus on seeking visual enhancement while ignore to boost the performance of the follow-up high-level vision tasks, \emph{e.g.,} object detection, depth estimation, and semantic segmentation. Hence, the generated fused images are difficult to be directly applied in intelligence systems with perception requirements. A completed pipeline of IVIF for practical applications is given in Figure~\ref{fig:pipeline}.
	
	In summary, the total goal of IVIF is to generate a fused image that not only can realize visual enhancement but also boost environment perception rate. 
	
	\begin{figure*}[!htb]
		\centering
		\setlength{\tabcolsep}{1pt} 
		\includegraphics[width=0.98\textwidth]{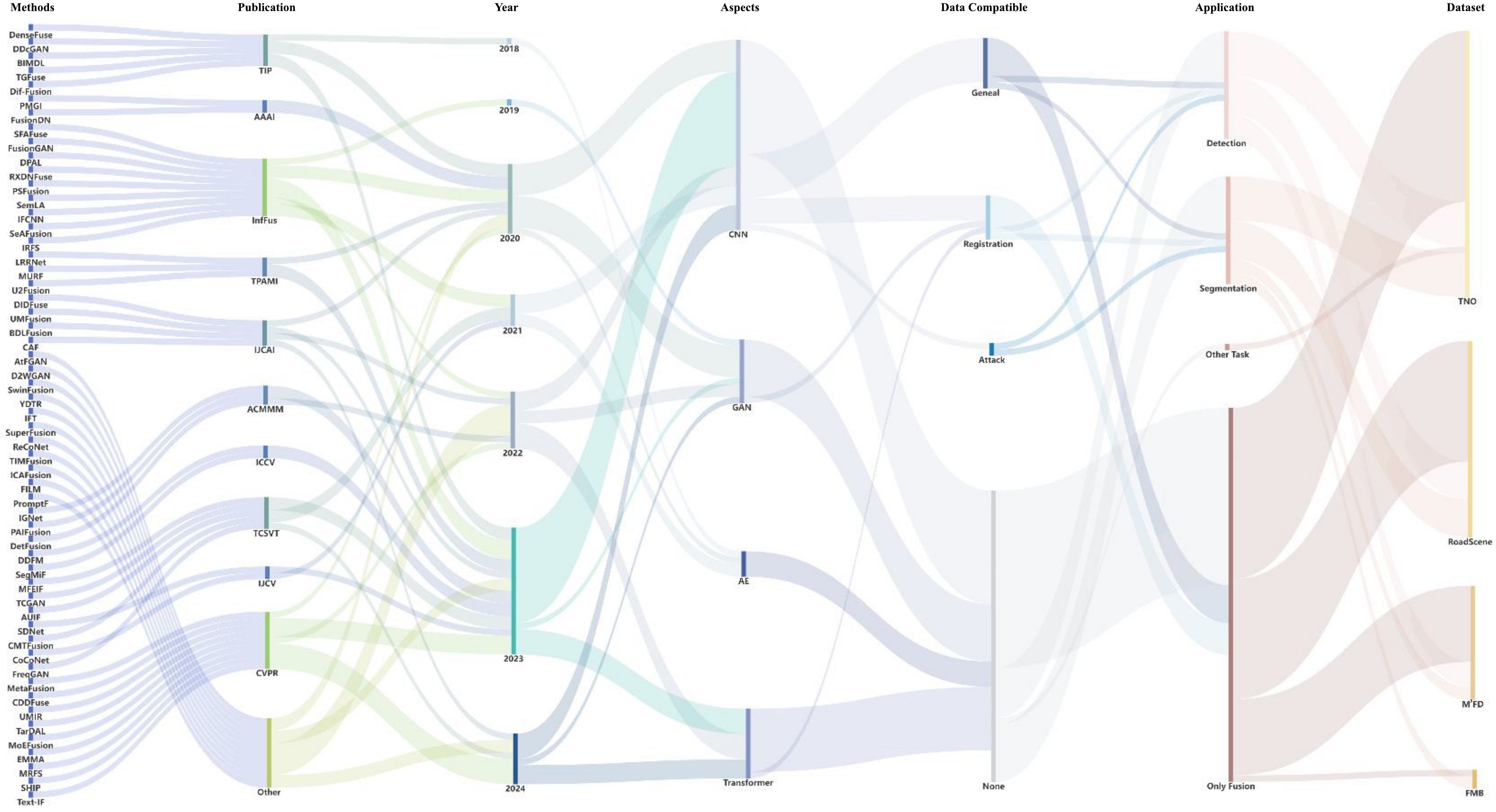}
		\caption{A classification sankey diagram containing typical fusion methods.}
		\label{fig: Dendrogram}
		\vspace{-0.4cm}
	\end{figure*}
	\vspace{-0.25cm}
	\section{Literature Survey}
	In this section, we first introduce a taxonomy for IVIF task from a multi-dimensional perspective, and then provide a detailed discussion in terms of widely used components of architectures and loss functions. The overview of our new taxonomy is given in Figure~\ref{fig: Dendrogram} and Table~\ref{tab:bigbiao}.
	\vspace{-0.25cm}
	\subsection{Fusion for Visual Enhancement}
	\subsubsection{AE-based Approaches}
\begin{figure*}[!htb]
		\centering
		\setlength{\tabcolsep}{1pt} 
		
		\includegraphics[width=0.98\textwidth]{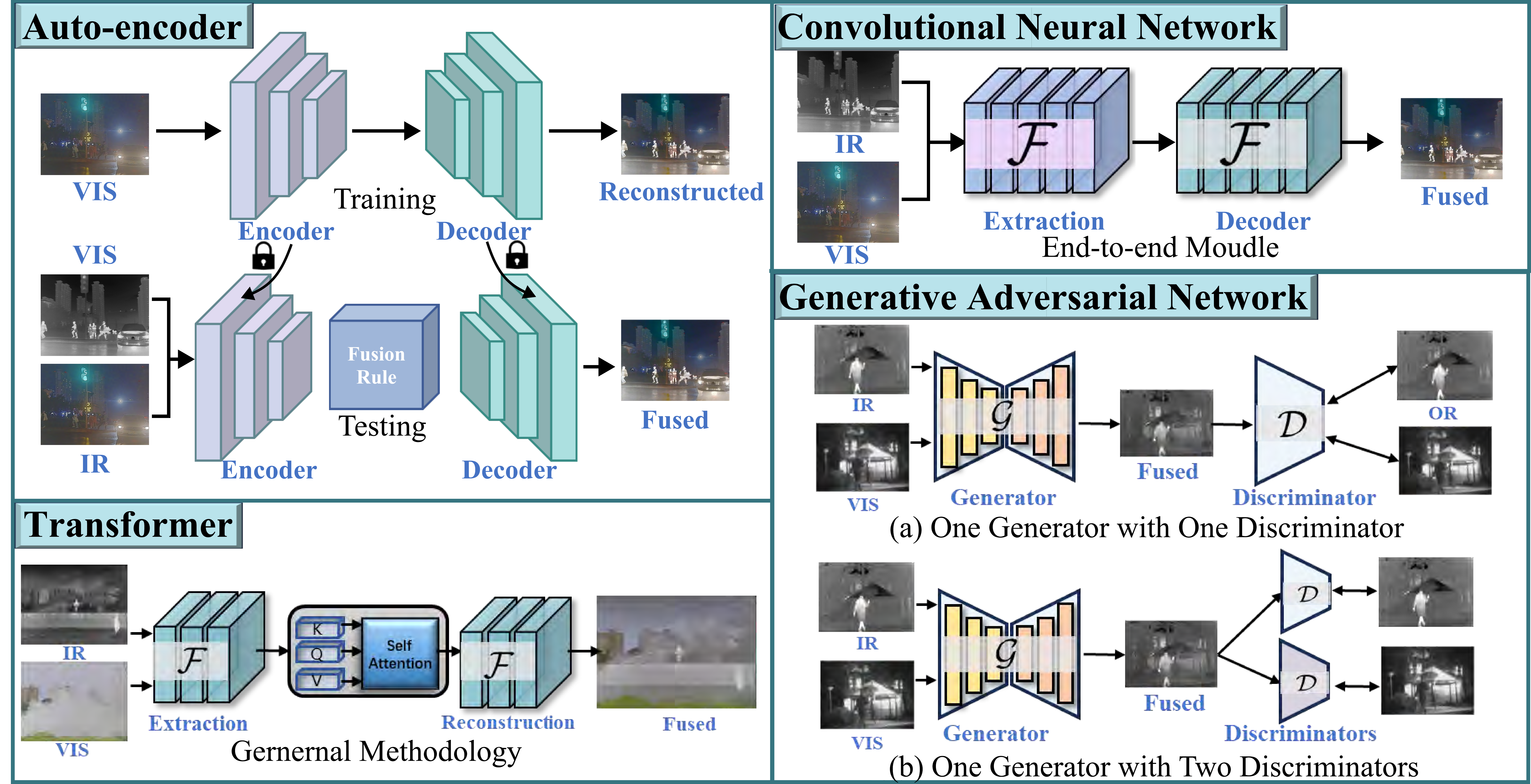}
		\vspace{-0.2cm}
		\caption{The basic phased processes of AE / CNN / GAN / Transformer-based IVIF methods.}
		\label{fig:111}
		\vspace{-0.5cm}
	\end{figure*}

	Auto-Encoder (AE)-based methods~\cite{raza2020pfaf, fu2021dual, jian2020sedrfuse, wang2022res2fusion, xu2021classification, zhu2021multiscale, fu2021effective, pan2021densenetfuse, zhao2021self} consist of two steps. First, an autoencoder is pre-trained using visible and/or infrared images. Second, the trained encoder is used for feature extraction, and the trained decoder is used for image reconstruction. The fusion between the encoder and decoder is usually performed according to manual fusion rules or is learned through a second training step using visible-infrared image pairs, as shown in Figure~\ref{fig:111}. Existing AE methods can be divided into two categories:
	
	i) Enhancements in fusion rules and data integration, aimed at improving multi-modality feature synthesis.
	
	{ii) Innovations in network architecture, which include new layer introductions and connection modifications.}

	\textbf{Fusion Rules.}~ DenseFuse~\cite{li2018densefuse}, a pioneer in fusion strategy methods, introduces two core approaches: an addition strategy for combining encoder-generated feature maps and an $l_1$-norm based strategy using softmax to select prominent features. These dual strategies yield superior over traditional methods. To further enhance feature integration and emphasize crucial information,  Li~\textit{et al.}~\cite{li2020nestfuse} innovate by merging a nested connection network with spatial/channel attention models for infrared-visible image fusion. It highlights spatial and channel importance via attention models, enhancing multi-scale deep feature fusion and surpassing other methods in objective and subjective evaluations.

	\textbf{Network Architecture.}~ 
	RFN-Nest~\cite{li2021rfn} distinguishes itself with a novel network architecture, introducing a Residual Fusion Network (RFN) that targets the enhancement of deep feature-based image fusion, focusing particularly on the sophistication of network design to preserve intricate details. Addressing the intricacies of infrared-visible image fusion, Liu~\textit{et al.}~\cite{liu2021learning} advance the field with its innovative network structure that incorporates multi-scale feature learning and an edge-guided attention mechanism, optimizing the architecture to enhance detail definition while reducing noise. In the realm of network innovation, Zhao~\textit{et al.}~\cite{zhao2020didfuse} employ an auto-encoder with a specialized design to segregate and integrate background and detail features effectively, emphasizing the strategic network modifications for thorough feature amalgamation. Extending the focus on network-driven solutions, Zhao~\textit{et al.}~\cite{zhao2021efficient} utilize algorithm unrolling to reconceptualize traditional optimization into a structured network process, meticulously engineered to segregate and fuse different frequency information, highlighting the strategic architectural advancements for improved fusion outcomes.
	
	\subsubsection{CNN-based Approaches}
	In CNN-based image fusion algorithms~\cite{liu2018infrared,liu2019infrared,hou2020vif,mustafa2020infrared,li2019infrared,li2022multilevel,an2020infrared,zhang2017infrared,feng2020fully,wang2019generative,zhu2022iplf}, the process generally involves three main steps: feature extraction, fusion, and image reconstruction, as shown in Figure~\ref{fig:111}. The key advantage of such algorithms lies in their ability to autonomously learn complex and high-level features from data. 
	
	In the realm of CNN-based methods for image fusion, three innovative approaches stand out:
	
	i) Optimization-inspired CNN methods leverage iterative integration and learnable modules for enhanced fusion efficiency.
	
	ii) Modifying loss functions crucially defines the unsupervised learning outcomes in IVIF tasks.
	
	iii) Architectural advancements focus on optimizing network designs, with NAS as a specialized approach for structural refinement.
	
	\textbf{Optimization-inspired.} Based on the natural priors of modal characteristics, optimization-model-inspired learning models are proposed for infrared and visible image fusion~\cite{li2023lrrnet,liu2020bilevel,zhao2021efficient}. These methods usually introduce the network into the iteration process, guided by the optimization objective, or  replace the numerical operations with learnable modules.
	Li~\textit{et al.}\cite{li2023lrrnet} introduce LRRNet, applying Low-Rank Representation to network design, thus enhancing interpretability. Liu~\textit{et al.}\cite{liu2020bilevel} present a bi-level optimization-based fusion method, focusing on image decomposition. Zhao~\textit{et al.}~\cite{zhao2021efficient} employ algorithm unrolling for image fusion, targeting low/high-frequency information across modalities.

	\textbf{Loss Functions}. Loss functions are crucial objectives, and for the unsupervised image processing task of IVIF, their setup is of significant importance. PIAFusion~\cite{Tang2022PIAFusion} employs an innovative illumination-aware loss, guided by a sub-network that assesses scene lighting, significantly improving image fusion under various lighting conditions. Expanding on context-aware processing, STDFusionNet~\cite{ma2021stdfusionnet} uses a loss function enhanced by a salient target mask to prioritize the integration of critical infrared features with visible textures, greatly improving feature integration and image clarity.

	\textbf{Structures.} Early works focusing on network structures concentrated on the utilization and fine-tuning of architectures like residual networks~\cite{li2021different, mustafa2020infrared, long2021rxdnfuse, xu2021drf, xu2022multi}, dense networks~\cite{li2020unsupervised,wang2021unfusion,ren2021infrared,long2021rxdnfuse,raza2021ir,liu2021two,xu2022cufd,yang2022tpfusion}, and U-Net, which will not be elaborated here. The methods introduced here, which focus on designing network structures, mostly incorporate novel technical approaches to upgrade fusion algorithms. 
	
	IGNet~\cite{ignet} innovatively combines CNN and Graph Neural Network (GNN) for infrared-visible image fusion. It starts with CNN-based multi-scale feature extraction, then employs a Graph Interaction Module to transform these features into graph structures for effective cross-modality fusion. Additionally, its leader node strategy in GCNs boosts information propagation, thereby preserving texture details more efficiently. {Yue~\textit{et al.} introduce Dif-Fusion~\cite{yue2023dif}, utilizing diffusion models to construct multi-channel  distributions, enabling direct generation of chromatic fused images with high color fidelity.}
	
	Specially, Neural Architecture Search (NAS) has achieved widespread developments for image fusion in recent years, which can automatically discover the desired architectures, avoiding massive handcrafted architecture engineering and dedicated adjustments. 
	As for the construction of super-net, SMoA~\cite{liu2021smoa} is proposed based on the auto-encoder paradigm, adequately representing the typical features based on two modality-specific encoders.
	To tackle blurred targets and detail loss, Liu~\textit{et al.}~\cite{liu2021searching} develop a hierarchically aggregated fusion method, aiming for comprehensive target and detail representation. Furthermore, Liu~\textit{et al.}~\cite{liu2022learn} propose a hardware latency-aware approach for crafting lightweight networks, reducing computational demands and aiding practical deployment.
	Recently, an implicitly-search strategy~\cite{liu2023task} is proposed with sufficient convergence of fusion, showing remarkable performances compared with existing  methods~\cite{liu2018darts,xu2019pc}.

	\subsubsection{GAN-based Approaches}
	Generative Adversarial Network (GAN)\cite{goodfellow2020generative} has demonstrated its effectiveness in modeling data distributions without label supervision. This unsupervised approach naturally suits the IVIF task, where GAN has become a major methodology. Existing methods can be classified into two categories, as shown in Figure~\ref{fig:111}:

	i) Single discrimination~\cite{xu2020lbp, xu2020infrared, wang2021new, yuan2020flgc, bhagat2020multimodal, ma2019fusiongan, ma2020ganmcc, li2019coupled, lebedev2019multisensor, gu2021advanced, hou2021generative} utilizes the original GAN to constrain the fused images as similar as one modality.
	
	ii) Dual discrimination~\cite{xu2019learning, ma2020ddcgan, li2020attentionfgan, zhou2021semantic, zhang2021gan, gao2022dcdr, rao2023gan, le2022uifgan, liu2022target, li2019infrared, li2020infrared, li2020multigrained, song2022triple, zhao2020fusion} utilizes two discriminators to balance the typical modality information.
	
	\textbf{Single discrimination.}  Ma~\textit{et al.} firstly propose the FusionGAN method~\cite{ma2019fusiongan}, which contains a generator, aiming to preserve the infrared intensities and textural details, and one discriminator, guaranteeing the textural details in the visible images. Subsequently, the same team~\cite{ma2020infrared} further improves the FusionGAN with two loss functions. The detail loss and edge loss are proposed to constrain the target boundaries of infrared targets for the FusionGAN.
	Amounts of methods focus on designing optimal generators to produce visual-appealing results. For instance, Fu~\textit{et al.}~\cite{fu2021image} apply densely connected blocks to construct the generator for preserving the textural details of fused images. Ma~\textit{et al.} propose GANMcC~\cite{ma2020ganmcc}, introducing a more balanced training with multi-classification for visible and infrared images. The generator consists of two sub-modules, including the gradient path for extracting texture details and the contrast path for  intensity information.

	\textbf{Dual discrimination.}  Xu~\textit{et al.} propose the DDcGAN~\cite{xu2019learning}, utilizing two discriminator to enforce the similarity with different modalities. Ma~\textit{et al.}  extend the DDcGAN by replacing the U-Net generator with densely connected blocks~\cite{ma2020ddcgan}. Following this paradigm, lots of methods are designed with one generator and two discriminators.  Li~\textit{et al.} propose AttentionFGAN~\cite{li2020attentionfgan}, introducing the multi-scale attention mechanisms into the construction of generator and discriminator. By the supervision of the attention loss, this method can keep the information of attention regions of source images. Zhou~\textit{et al.} propose the SDDGAN~\cite{zhou2021semantic}, designing information quantity discrimination to estimate the informative richness. Gao~\textit{et al.}~\cite{gao2022dcdr} construct the generator with dense connection and multi-scale fusion and propose the reconstruction losses of content and modality.
	Rao~\textit{et al.}~\cite{rao2023gan}  present the attention and transition modules to composite the generator, to filter out the noise and enhance the quality in adverse conditions. {Wang~\textit{et al.}~\cite{wang2022infrared} utilizes dual discriminators in ICAFusion to ensure balanced fusion by matching the distribution of fused results with source images, while interactive attention modules enhance feature extraction and reconstruction. Further, they developed FreqGAN~\cite{wang2024freqgan}, incorporating dual frequency-constrained discriminators that dynamically adjust the weights for each frequency band. }

	The main drawback of single discrimination is that most of methods only provide one single modality supervision, making the fused images similar either to visible modality or infrared modality (\textit{i.e.,} modality unbalance), which causes the complementary features (\textit{i.e.,} pixel intensities of infrared images and texture details of visible images) cannot
	be preserved simultaneously. Therefore dual discrimination for diverse modality are introduced to address this issue. However, the major obstacle of dual discrimination is to guide modality-specific discriminator to extract diverse modality characteristics including the salient objects from thermal radiation and rich textural details from visible modality.
	
	\subsubsection{Transformer-based Approaches}
	Benefiting from the strong representation capabilities and long-range dependencies based on self-attention mechanisms, researchers are looking at ways to leverage this mechanism into infrared-visible image fusion~\cite{zhao2021dndt, li2022cgtf, liu2022mfst, yang2023dglt, qu2022transfuse, tang2022matr}. 	Transformer-based fusion methods mostly utilize the combined CNN-transformer network, aggregating the CNN blocks to extract the shallow features and leveraging the transformer blocks to construct the long-range dependence.

	VS~\textit{et al.}~\cite{vs2022image} pioneer the use of Transformers for image fusion, introducing a multi-scale auto-encoder and a spatio-transformer fusion strategy to aggregate local and global feature information.
	Ma~\textit{et al.}~\cite{ma2022swinfusion} present the SwinFusion, leveraging the swin transformer~\cite{liu2021swin} to construct two kinds of attention modules. The self and cross-attention mechanisms are proposed to integrate the long-range dependency between specific modalities and cross-domain.  Tang~\textit{et al.} present a y-shape transformer YDTR~\cite{tang2022ydtr} to construct the encoding and decoding branches by the combination of CNN and transformers, cascading vision transformer blocks~\cite{yin2022vit}. Furthermore,  Tang~\textit{et al.}~\cite{tang2023tccfusion} presents the local-global parallel architecture, consisting of local feature extraction and global feature extraction modules to better describe   local complementary features with local complementary features. Lately, DATfusion~\cite{tang2023datfuse} has been proposed to realize the image fusion via the interaction of dual attention and transformer block, which plays a vital role in preserving vital features and  global information preservation.
	\{CMTFusion~\cite{park2023cross} introduces cross-modal transformers to efficiently retain complementary information by removing spatial and channel redundancies, with a gated bottleneck enhancing cross-domain interactions for improved feature fusion.
	Yi~\textit{et al.}~\cite{yi2024text} apply a Transformer-based model Text-IF for text-guided fusion, addressing degradation and enabling interactive outcomes by leveraging a text semantic encoder and semantic interaction fusion decoder.
	Zhao~\textit{et al.}~\cite{zhao2023cddfuse} introduce CDDFuse, a dual-branch Transformer-CNN framework for IVIF that uses decomposition to extract cross-modality features and hybrid modules from Restormer\cite{zamir2022restormer} for lossless information transmission. 
    Liu~\textit{et al.}~\cite{liu2024promptfusion} leverages prompt-based learning from Vision-Language Models to guide image fusion, enhancing target identification and fusion quality through semantic prompts.
	
	{Transformer-based approaches effectively capture long-range dependencies, crucial for understanding cross-modality relationships, while self-attention helps retain global context in image fusion. However, these methods demand high computational resources and large memory, making real-world deployment challenging.}
	
	\begin{figure}[!htb]
		\centering
		\setlength{\tabcolsep}{1pt} 
		
		\includegraphics[width=0.48\textwidth]{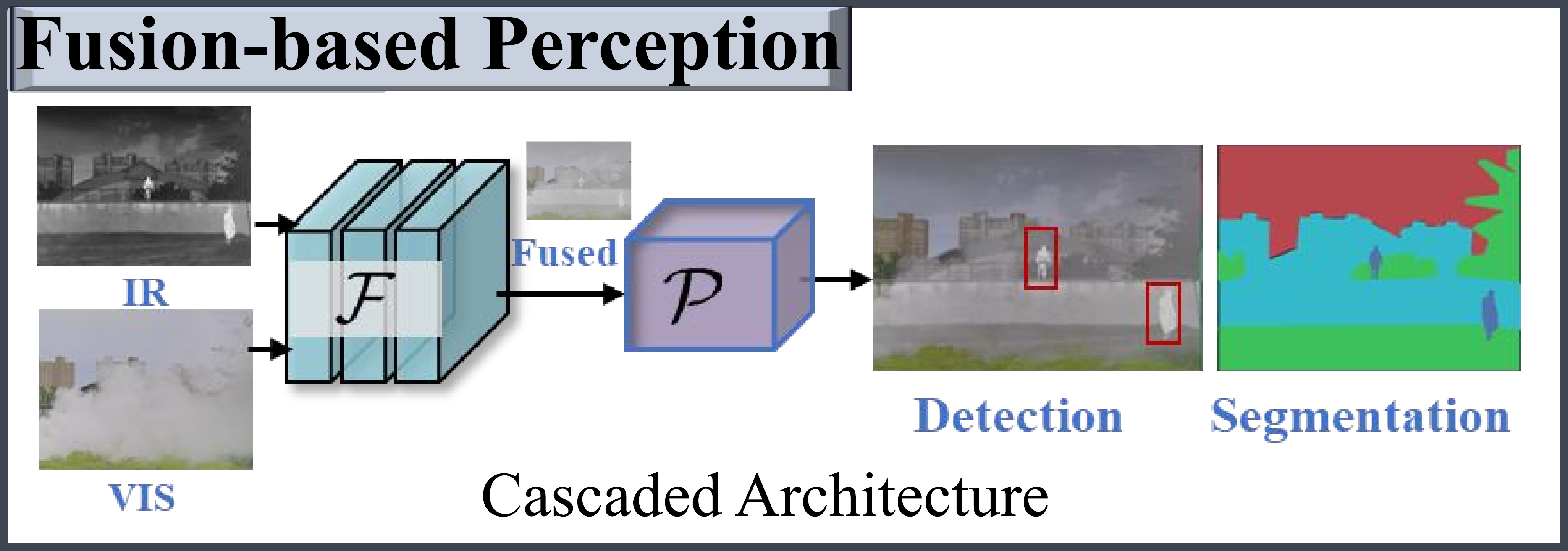}
		\vspace{-0.2cm}
		\caption{{The basic phased processes of application-oriented IVIF methods.}}
		\label{fig:222}
		\vspace{-0.85cm}
	\end{figure}
	
	\subsection{Application-oriented}
	{Multi-modality infrared and visible image fusion has gained significant attention due to its broad range of real-world applications (\textit{e.g.,} automatic driving and robotic operations). 
	Among all the tasks, object detection and semantic segmentation have become particularly central, as they are the most relevant tasks connected to fusion-related applications, as shown in Figure~\ref{fig:222}. Numerous methods and datasets have been proposed to tackle these challenges, making them critical topics of interest and key directions for future research.}

	{In this part, we will delve into the methods specifically developed for these two tasks and offer a detailed exploration of their recent progress, while also introducing other relevant tasks within this domain.}

	\subsubsection{Object detection}
	
	{As for object detection, Liu~\textit{et al.} pioneered the exploration of image fusion and object detection, termed TarDAL~\cite{liu2022target} with the largest multi-modality object detection datasets M$^3$FD.  TarDAL proposed a bi-level optimization formulation to model the inherent relationship between both tasks and unroll the optimization for a target-ware bi-level learning network.
	Zhao~\textit{et al.}~\cite{zhao2023metafusion} introduce the meta-feature embedding to achieve the compatibility between object detection and image fusion. 
	Sun~\textit{et al.} present the impressive DetFusion~\cite{sun2022detfusion}, leveraging the detection-driven information to guide the optimization of fusion by shared attention mechanisms. The object-aware loss also plays a key role in learning the pixel-level information from object locations. Cao~\textit{et al.} propose the MoE-Fusion~\cite{cao2023multi}, integrating a mixture of local-global experts to dynamically extract effective features of respective modalities. This dynamic feature learning for local information and global contrast demonstrates the effectiveness of object detection. Zhao~\textit{et al.} propose MetaFusion~\cite{zhao2023metafusion}, which leverages meta-feature embedding from object detection to align semantic and fusion features, enabling effective joint learning.}
	
	{In contrast to fusion-based methods, several approaches focus on object detection using infrared and visible images without producing fused images. These methods enhance detection by leveraging cross-modality interactions through attention mechanisms, iterative refinement of spectral features, probabilistic ensembling of detections, and aligning features between modalities for improved precision~\cite{chen2022multimodal, zhang2020multispectral, zhang2019weakly, zhang2019cross}.}
	
	\subsubsection{Semantic Segmentation}
	
	For semantic segmentation, SeAFusion~\cite{tang2022image} cascades image fusion with segmentation tasks, introducing semantic loss to enhance the information richness of fusion through loop-based training. Tang \textit{et al.} propose SuperFusion~\cite{SuperFusion_22}, a versatile framework for multi-modality image registration, fusion, and semantic perception. PSFusion~\cite{tang2023rethinking} introduces progressive semantic injection at the feature level, considering the semantic needs for fusion. It also shows that with fewer computational resources, image-level fusion provides comparable performance to feature fusion for perception tasks.
	{Lately, Liu~\textit{et al.} proposed SegMiF~\cite{liu2023multi}, which leverages dual-task correlation to enhance both segmentation and fusion performance, introducing hierarchical interactive attention for fine-grained task mapping and collecting the largest full-time benchmark for these tasks. Zhang~\textit{et al.} proposes MRFS~\cite{zhang2024mrfs}, a coupled learning framework that integrates image fusion and segmentation through mutual reinforcement, achieving enhanced visual quality and more accurate segmentation results.}
	
	{In addition to other fusion-based segmentation methods, two-stream infrared-visible semantic segmentation has gained traction. These approaches typically fuse modality-specific features at either the encoder or decoder stage. Encoder-based fusion methods aggregate features early, as shown in works like~\cite{sun2019rtfnet, zhang2023cmx, zhang2021abmdrnet, zhao2023mitigating}, while decoder-based fusion combines features during reconstruction, as demonstrated in~\cite{ha2017mfnet, zhou2021gmnet, zhou2022edge, li2022rgb}. Together, these two-stream methods complement the broader landscape of fusion-based semantic segmentation techniques.}

	Recently, uniform solutions to bridge image fusion and semantic perception have become an attractive topic.  Liu~\textit{et al.} \cite{liu2023bi} introduce the unified bi-level dynamic learning paradigm to guarantee image fusion that has visually appealing results and can serve downstream perception tasks.  {Liu~\textit{et al.}~\cite{CAF} use detection and segmentation tasks to guide the automated search process of the loss function, freeing up manual effort and enabling the construction of fusion methods for perception tasks in the CAF framework. TIMFusion~\cite{liu2023task} is proposed to discover hardware-sensitive networks through implicit architecture search, achieving fast adaptation for diverse tasks via pretext meta-initialization, while enhancing visual quality and supporting various semantic perception tasks.}

	\subsubsection{Other perception tasks}
	
	{Infrared and visible modalities are not only useful for fusion tasks but also play a key role in various other perception applications, such as object tracking, crowd counting, salient object detection, and depth estimation.}
	
	{\textbf{Object Tracking}~\cite{cvejic2007effect, zhang2021siamcda, li2020challenge, xiao2022attribute, zhu2021rgbt, zhang2019multi, tang2022temporal, feng2020learning, tang2023exploring} utilizes pixel-wise, feature-level, and decision-based fusion of RGB-T modalities to enhance tracking robustness and reliability.}
	{\textbf{Crowd Counting}~\cite{liu2021cross, tang2022tafnet, zhou2022defnet, zhou2023mc, wu2022multimodal, liu2023rgb, yang2024cagnet} benefits from multi-modality feature fusion to improve density estimation and crowd prediction.}
	{\textbf{Salient Object Detection}~\cite{wang2023interactively, zhang2019rgb, tu2021multi, liao2022cross, huo2021efficient} leverages two-stream frameworks and fusion modules to combine complementary cues from RGB-T data for more accurate object delineation.}
	{\textbf{Depth Estimation}~\cite{liu2021searching, park2023cross, zhi2018deep, liang2022deep, guo2023unsupervised} is enhanced by spectral transfer and style transfer methods to handle varying illumination conditions and improve cross-modality matching.}

	\begin{figure*}[!htb]
		\centering
		\setlength{\tabcolsep}{1pt} 
		
		\includegraphics[width=0.98\textwidth]{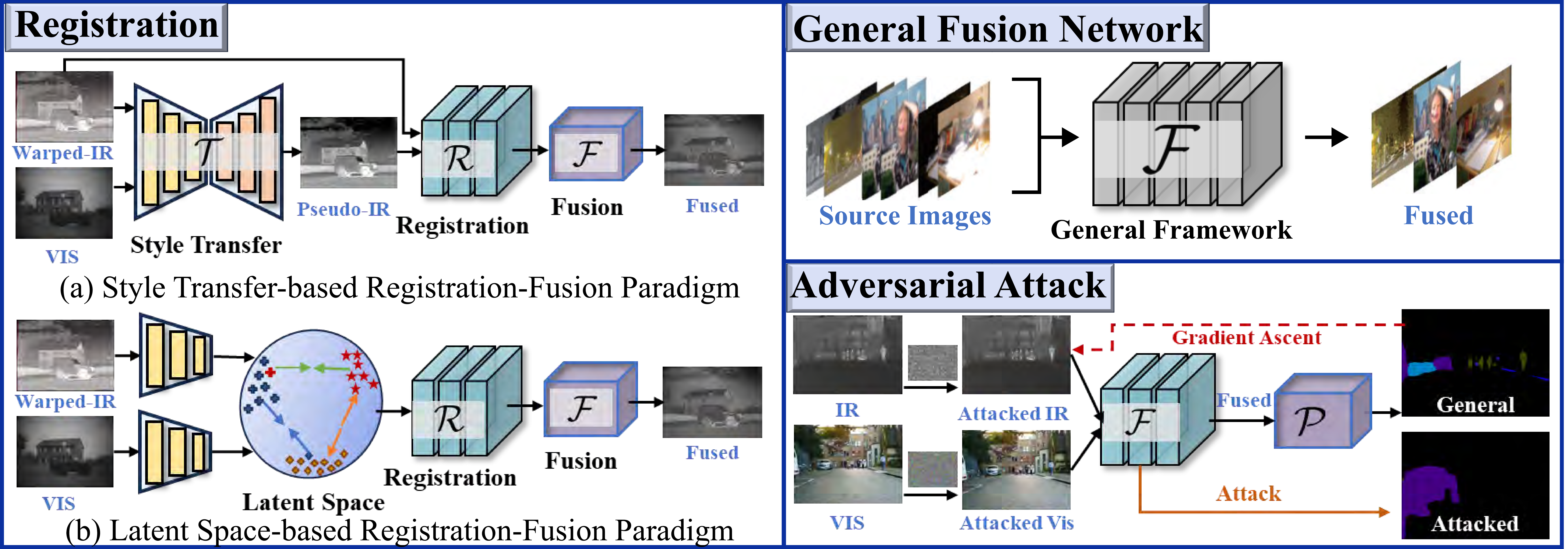}
		\vspace{-0.2cm}
		\caption{The basic phased processes of data compatible IVIF methods.}
		\label{fig:333}
		\vspace{-0.5cm}
	\end{figure*}

	\subsection{Data Compatible}
	\subsubsection{Registration-free approaches}
	A plethora of fusion methods tailored for well-aligned multi-modality images have emerged, while fusion methods designed for imperfectly aligned multi-modality images have just started to draw attention. Existing fusion methods for imperfectly aligned multi-modality images can be divided into two categories:
	
	i) Generation of pseudo labels, which transforms multi-modality registration into single-modality registration, called Style Transfer-based Methods.
	
	ii) Construction of a modality-independent feature space, where multi-modality image features are mapped to a shared space and shared features are utilized to predict deformation fields, called Latent Space-based Methods.

	\textbf{Style Transfer-based Methods.} A general scheme involves the joint learning of the Modality-Transfer Network (MTN) and Spatial-Transformation Network (STN). Due to the lack of direct supervision between cross-modality images, existing methods~\cite{Nemar_20, UMF, RFNet_22} utilize MTN to transform images from one modality to another, thereby generating corresponding pseudo-labels. Subsequently, STN is employed to predict the spatial displacements between the source image and the pseudo-labeled image.
	
	Nemar~\cite{Nemar_20} pioneers the use of mono-modality metrics for training multi-modality image registration. It utilizes a bidirectional training approach, with options for ``spatial registration first, image translation later" and ``image translation first, spatial registration later", which encourages the image translation network to generate pseudo-labeled images with preserved geometric properties, leading to improved registration.
	 
	UMFusion~\cite{UMF} proposes a highly robust unsupervised framework for infrared and visible image fusion, focusing on alleviating ghosting artifacts caused by misaligned multi-modality images. Specifically, a cross-modality generation-registration paradigm is introduced to generate pseudo labels, aiming to reduce large modality discrepancies between infrared and visible images from pixel level.
	
	With similar goals, RFNet~\cite{RFNet_22} achieves the registration and fusion of multi-modality images through mutual reinforcement, rather than treating them as independent optimization objectives. In RFNet, multi-modality image registration is defined as a coarse-to-fine process, and fine registration and fusion are synergistically combined through interactive learning to improve the quality of fused images and strengthen the reciprocal promotion effect of fusion on registration.

	\textbf{Latent Space-based Methods.} The main idea is to extract common features from multi-modality images, which reduces modality discrepancies between infrared and visible images from the feature level. Typically, MURF~\cite{MURF_23} argues that modality-independent features are crucial for registration, and therefore proposes to map multi-modality features into a modality-independent shared space. MURF divides the registration into two stages, i.e., coarse registration and fine registration. In the coarse registration stage, contrastive learning is used to constrain the generation of shared feature representations. Two other methods, SuperFusion~\cite{SuperFusion_22} and ReCoNet~\cite{ReCoNet_22}, learn deformation parameters for registration directly from cross-modality features. The former corrects the geometric distortion of the inputs by estimating the bidirectional deformation fields under the supervision of photometric and endpoint constraints and symmetrically combines registration and fusion to achieve mutual promotion. The latter develops a recurrent correction network to explicitly compensate for geometric distortions, which in turn alleviates ghosts in fused images.

	\begin{table*}[]
		\caption{AN OVERVIEW OF REPRESENTATIVE DEEP LEARNING-BASED IVIF METHODS.}
		\renewcommand\arraystretch{1.1} 
		\setlength{\tabcolsep}{2.3mm}
		\begin{tabular}{|c|l|l|l|l|l|l|l|}
			\hline
			& Aspects              & Methods   & Publication&F&D&T             & Core Ideas                                  \\
			\hline
			\multirow{34}{*}{\adjustbox{angle=90}{\textbf{\textcolor{cyan}{Fusion for Visual Enhancement}}}} & \cellcolor{cyan!10}& DenseFuse~\cite{li2018densefuse} &TIP & \checkmark&-&-&Dense connection feature extraction with $l1$-norm fusion rule\\
			&\cellcolor{cyan!10}& SEDRFuse~\cite{jian2020sedrfuse} & TIM & \checkmark &-&-&Symmetric residual block feature extraction with attention-guided fusion rule \\
			&\cellcolor{cyan!10}& DIDFuse~\cite{zhao2020didfuse} & IJCAI & \checkmark &\checkmark&-&Proposing a data-driven auto-encoder based feature decomposition network\\
			&\cellcolor{cyan!10}& MFEIF~\cite{liu2021learning} & TCSVT & \checkmark&-&-& Multi-scale feature extraction with edge attention fusion rule\\
			&\cellcolor{cyan!10}{\shortstack[l]{Auto-Encoder}}& RFN-Nest~\cite{li2021rfn} & TIM & \checkmark &-&\checkmark&Residual fusion network with learnable fusion rule\\
			&\cellcolor{cyan!10}&Re2Fusion~\cite{wang2022res2fusion}&TIM& \checkmark &-&-& Dense Residual structure with double no-nlocal attention fusion models \\
			&\cellcolor{cyan!10}&SMoA~\cite{liu2021smoa}&SPL& \checkmark &-&-& Automatic feature extractor with salient weight based fusion rule\\
			&\cellcolor{cyan!10}&SFAFuse~\cite{zhao2021self}&InfFus& \checkmark &-&-&Extract features
			with adaption via a self-supervised strategy\\
			
			\cline{2-8}
			\hhline{~-------} 
			
			& \cellcolor{cyan!13} & FusionGAN~\cite{ma2019fusiongan} & InfFus& \checkmark &-&-&Signal discriminator to keep both the thermal radiation and the texture details \\
			&\cellcolor{cyan!13}&TCGAN~\cite{zhang2023transformer}&TMM& \checkmark &-&-&Transformer-based conditional GAN for prior knowledge integration\\
			&\cellcolor{cyan!13}&DPAL~\cite{ma2020infrared}&InfFus& \checkmark &-&-&Designing  detail loss and target edge-enhancement loss to improve the quality\\
			&\cellcolor{cyan!13} &AtFGAN~\cite{li2020attentionfgan}&TMM& \checkmark &-&-&Integrate multi-scale attention
			mechanism into both generator and discriminator\\
			&\cellcolor{cyan!13}GAN &D2WGAN~\cite{li2020infrared}&InfSci& \checkmark &-&-&Keeping pixel intensity and texture information by dual wasserstein discriminators\\

			&\cellcolor{cyan!13}&GANMcC~\cite{ma2020ganmcc}&TIM& \checkmark &-&-&GAN
			with multi-classification constraints for addressing unbalanced fusion\\
			&\cellcolor{cyan!13}& DDcGAN~\cite{ma2020ddcgan} & TIP & \checkmark&\checkmark&-&Dual discriminators for main the details/content from visible/infrared image\\
			&\cellcolor{cyan!13}& {ICAFusion}~\cite{wang2022infrared} & TMM & \checkmark&\checkmark&-&Dual cross-attention feature fusion with iterative interaction mechanism\\
			&\cellcolor{cyan!13}& {FreqGAN}~\cite{wang2024freqgan} & TCSVT & \checkmark&\checkmark&-&Wavelet-based decomposition with feature aggregation for detail preservation\\
			
			\cline{2-8}
			\hhline{~-------}
			
			&\cellcolor{cyan!16}& RXDNFuse~\cite{long2021rxdnfuse}	&InfFus& \checkmark &-&-&Aggregated residual dense network for overcoming the manual fusion rule \\
			&\cellcolor{cyan!16}&STDFusionNet~\cite{ma2021stdfusionnet}&TIM& \checkmark &-&-& Salient target detection weight map in the loss function\\
			&\cellcolor{cyan!16}& MetaFusion~\cite{zhao2023metafusion} & CVPR & \checkmark&-&-&Meta-feature embedding
			model is designed to generate object semantic features\\
			&\cellcolor{cyan!16}&BIMDL~\cite{liu2020bilevel} &TIP& \checkmark &\checkmark&-&Layer-guided bilevel optimization modeling with adaptive weight integration\\
			&\cellcolor{cyan!16}&AUIF~\cite{zhao2021efficient} &TCSVT& \checkmark &\checkmark&-&Presenting an algorithm unrolling based interpretable fusion network\\
			&\cellcolor{cyan!16} & LRRNet~\cite{li2023lrrnet} & TPAMI & \checkmark &-&\checkmark&Designing learnable representation model with detail-to-semantic loss function\\
			&\cellcolor{cyan!16}CNN&MgAN-Fuse~\cite{li2020multigrained}&TIM& \checkmark &-&-&Implanting multigrained attention to preserve the foreground target/context\\
			&\cellcolor{cyan!16}&CUFD~\cite{xu2022cufd}&CVIU& \checkmark &-&-&Extracting shallow and deep features a with different emphases\\
			&\cellcolor{cyan!16}& IGNet~\cite{ignet} &ACMMM & \checkmark &-&\checkmark&Building cross-modality relationship via graph neural networks\\
			&\cellcolor{cyan!16}& PSFusion~\cite{tang2023rethinking} & InfFus & \checkmark &-&-& Designing progressive semantic injection and scene fidelity constraints\\
			&\cellcolor{cyan!16}& {Dif-Fusion}~\cite{yue2023dif} & TIP & \checkmark &-&-& Diffusion-based multi-channel fusion for better color fidelity and detail retention\\

			\cline{2-8}
			\hhline{~-------}
			
			&\cellcolor{cyan!19}& SwinFusion~\cite{ma2022swinfusion} & JAS & \checkmark& \checkmark&\checkmark&Integration of complementary information and global interaction via attention\\
			&\cellcolor{cyan!19}& YDTR~\cite{tang2022ydtr} & TMM & \checkmark &-&-&Acquiring the local/context information by a dynamic  transformer module\\
			&\cellcolor{cyan!19}& IFT~\cite{vs2022image} & ICIP & \checkmark  &-&-&Developing a transformer-based rule to attends local and long-range information \\
			&\cellcolor{cyan!19}& CDDFuse~\cite{zhao2023cddfuse} & CVPR & \checkmark &\checkmark&\checkmark& Proposing a two-stream correlation-driven feature decomposition network\\
			&\cellcolor{cyan!19}Transformer & TGFuse~\cite{rao2023tgfuse} & TIP & \checkmark &-&-& Learning the global fusion relations and reflecting the specific characteristics\\
			&\cellcolor{cyan!19}& {CMTFusion}~\cite{park2023cross} & TCSVT & \checkmark &-&-& Gated bottleneck to integrate cross-domain interactions for source images\\
			&\cellcolor{cyan!19}& {Text-IF}~\cite{yi2024text} & CVPR & \checkmark &-&-& Text-guided framework for degradation-aware and interactive processing\\
			&\cellcolor{cyan!19}& {PromptF}~\cite{liu2024promptfusion} & JAS & \checkmark &-&-& Harmonized Semantic Prompt Learning for improved detail and target extraction\\
			
			\hline
			\multirow{18}{*}{\adjustbox{angle=90}{\textbf{\textcolor{gray}{Data Compatible}}}} &\cellcolor{gray!10}
			
			&GCRF~\cite{luo2023general}&SR&-&\checkmark&-&Constructing a general framework to explores generation registration pattern \\
			&\cellcolor{gray!10}&UMIR~\cite{Nemar_20}&CVPR&-&\checkmark&-&Alleviating manual similarity measure by translation image-to-image network\\
			&\cellcolor{gray!10}& UMFusion~\cite{UMF} &IJCAI&\checkmark&\checkmark&-& Style Transfer-based cross-modality generation-registration paradigm for fusion\\
			& \cellcolor{gray!10}Registration & MURF~\cite{MURF_23} & TPAMI&\checkmark&\checkmark&-&Mutually reinforced registration and fusion via three well deigned module\\
			&\cellcolor{gray!10}& SuperFusion~\cite{SuperFusion_22}& JAS&\checkmark&\checkmark&\checkmark&Estimating bidirectional deformation fields and integrate semantic constraint\\
			&\cellcolor{gray!10}& ReCoNet~\cite{ReCoNet_22}&ECCV &\checkmark&\checkmark&-&Developing a recurrent light
			network that corrects distortions and artifacts\\
			&\cellcolor{gray!10}& SemLA~\cite{SemLA_23}&InfFus &\checkmark&\checkmark&-&Explicitly embedding semantic information at all stages of the network\\
			\cline{2-8}
			\hhline{~-------}
			& \cellcolor{gray!15}Attack
			& PAIFusion~\cite{liu2023paif} & ACMMM &\checkmark&-&\checkmark&A  perception-aware network to improve robustness under adversarial attack\\
			
			\cline{2-8}
			\hhline{~-------}
			
			&\cellcolor{gray!20}
			&U2Fusion~\cite{xu2020u2fusion}&TPAMI&\checkmark&\checkmark&-&Feature measurement that automatically estimates corresponding source images\\
			&\cellcolor{gray!20}&SDNet~\cite{zhang2021sdnet}&IJCV&\checkmark&\checkmark&-&Universal loss with squeeze and decomposition  for multiple fusion tasks\\
			&\cellcolor{gray!20}&IFCNN~\cite{zhang2020ifcnn}&InfFus&\checkmark&\checkmark&-&Extracting the salient features with appropriate fusion\\
			&\cellcolor{gray!20}&PMGI~\cite{zhang2020rethinking}&AAAI&\checkmark&\checkmark&-&Uniform loss function that divides the network into gradient and intensity\\
			&\cellcolor{gray!20}General&FusionDN~\cite{xu2020fusiondn}&AAAI&\checkmark&\checkmark&-&IQA-based loss functions for general image fusion tasks\\
			&\cellcolor{gray!20}&CoCoNet~\cite{liu2023coconet}&IJCV&\checkmark&\checkmark&\checkmark&Couple contrastive constrains with adaptive loss function ensemble\\
			&\cellcolor{gray!20}&DDFM~\cite{zhao2023ddfm}&ICCV&\checkmark&\checkmark&\checkmark& Denoising diffusion sampling model and Bayesian rectification\\
			&\cellcolor{gray!20}& {FILM}~\cite{zhaoimage} & ICML & \checkmark &-&-& Introducing a semantic prompt network based on vision-language models\\
			&\cellcolor{gray!20}& {EMMA}~\cite{zhao2024equivariant} & CVPR & \checkmark &-&-& Equivariant self-supervised learning with pseudo-sensing for feature consistency\\
			
			\hline
			\multirow{11}{*}{\adjustbox{angle=90}{\textbf{{Application-oriented}}}} & \cellcolor{blue!20}
			&TarDAL~\cite{liu2022target}&CVPR&\checkmark&-&\checkmark&Seeking common/unique modality features and joint learning by loss function\\
			&\cellcolor{blue!20}&DetFusion~\cite{sun2022detfusion}&ACMMM&\checkmark&-&\checkmark&Utilizing object-related information to guide the fusion process\\
			&\cellcolor{blue!20}&SeAFusion~\cite{tang2022image}&InfFus&\checkmark&-&\checkmark&Cascading the fusion and segmentation module and leveraging the semantic loss\\
			&\cellcolor{blue!20}&SegMiF~\cite{liu2023multi}&ICCV&\checkmark&-&\checkmark&Joint learning fusion and segmentation features via cross modality attention\\
			&\cellcolor{blue!20}Perception&BDLFusion~\cite{liu2023bi}&IJCAI&\checkmark&-&\checkmark&Improving visual quality and semantic information via bi-level optimization\\
			&\cellcolor{blue!20}&IRFS~\cite{wang2023interactively}&InfFus&\checkmark&-&\checkmark&Interactively reinforced multi-task framework to bridge  fusion and SOD\\
			&\cellcolor{blue!20}&MoE-Fusion~\cite{cao2023multi}&ICCV&\checkmark&-&\checkmark&Dynamic framework with a mixture experts for prevent features lose
			\\
			&\cellcolor{blue!20}&TIMFusion~\cite{liu2023task}&TPAMI&\checkmark&-&\checkmark& Implicit architecture search and support  adaptation via meta initialization\\
			&\cellcolor{blue!20}&{MRFS}~\cite{zhang2024mrfs}&CVPR&\checkmark&-&\checkmark& Coupled learning framework for enhancing fusion and segmentation\\
			&\cellcolor{blue!20}&{MetaFusion}~\cite{zhao2023metafusion}&CVPR&\checkmark&-&\checkmark& Introducing a meta-feature embedding model with mutual promotion learning \\
			&\cellcolor{blue!20}&{CAF}~\cite{CAF}&IJCAI&\checkmark&-&\checkmark& Automatic loss function design optimized by perception tasks.\\
			\hline
			
		\end{tabular}
		\label{tab:bigbiao}
	\end{table*}
	
	\subsubsection{General fusion approach}
	
	In multi-modality image fusion research, a general fusion framework serves as a vital technique for integrating different imaging technologies. These frameworks demonstrate potential applications in fields such as medical imaging and infrared and visible imaging, due to their algorithmic generality and superior scalability. They accomplish comprehensive feature extraction, from textural details to high-level semantic information, through in-depth feature learning and structural optimization. Moreover, these frameworks make use of diverse loss functions, like perceptual and structural similarity loss, to ensure the quality and integrity of the fused images. To address the core aspects of general fusion frameworks in multi-modality image fusion:
	
	i) Loss Function Enhancements: Innovations focus on advanced loss metrics to enhance image quality.
	
	ii) Architectural Innovations: Upgrades target better feature extraction and network efficiency.
	
	\textbf{Loss function} innovation is pivotal for improving fusion model performance. Key developments include:
	Zhang \textit{et al.}~\cite{zhang2020ifcnn} pioneer with a perceptual loss-trained convolutional framework, achieving detailed enhancement and broad task applicability without post-processing. Xu \textit{et al.}~\cite{xu2020u2fusion} innovatively use information theory for input weighting, enhancing loss optimization and task unification. Xu \textit{et al.}~\cite{xu2020fusiondn} employ NR-IQA and entropy for weight adjustment, ensuring stable quality and reducing ground truth reliance. Lastly, Liu \textit{et al.}~\cite{liu2023coconet} leverage contrastive learning for loss refinement, boosting cross-modal consistency and information retention. {Zhao~\textit{et al.} propose EMMA~\cite{zhao2024equivariant}, a self-supervised learning paradigm that refines conventional fusion loss by incorporating a pseudo-sensing module and sensing loss, effectively simulating the perceptual imaging process.}

	\textbf{Structure.} The structural evolution is characterized by innovative designs tailored for multi-modal data integration:
	Li \textit{et al.}~\cite{li2021multiple} introduce a framework with multiple task-specific encoders and a universal decoder, enhancing the precision of feature extraction and facilitating knowledge exchange across various fusion tasks. Liu \textit{et al.}~\cite{liu2023task} leverage Implicit Architecture Search (IAS) to dynamically optimize the model structure, incorporating insights from related perception tasks to improve unsupervised learning and model versatility. Liu \textit{et al.}~\cite{liu2020bilevel} advance structural optimization in multi-modal fusion by segregating base and detail layers processing, employing a bilevel optimization to refine texture and detail representation while ensuring structural integrity in the fused output. {Zhao~\textit{et al.} propose FILM~\cite{zhaoimage}, a method leverages large language models by generating semantic prompts from images, using ChatGPT to produce textual descriptions that guide the fusion process through cross-attention, enhancing both feature extraction and contextual understanding.}

	\subsubsection{Attack} Adversarial attacks~\cite{wei2022simultaneously, zhang2022revisiting}, which add indistinguishable perturbations to images, are easy to fool the estimation of neural networks. The basic workflow is plotted in Figure~\ref{fig:333}.
	The vulnerability of networks for multi-modality vision under adversarial attacks has not been widely investigated. 
	Considering the robustness of multi-modality segmentation against adversarial attacks, PAIFusion~\cite{liu2023paif} is the first to leverage image fusion to enhance robustness. This work identifies fragile fusion operations and rules through detailed analysis. In our view, this is an urgent and challenging topic for future research, as it plays a crucial role in ensuring the robustness and safety of real-world applications.
	\vspace{-0.3cm}

		\begin{figure}[t]
		
		\centering
		\setlength{\tabcolsep}{1pt} 
		
		\includegraphics[width=0.45\textwidth]{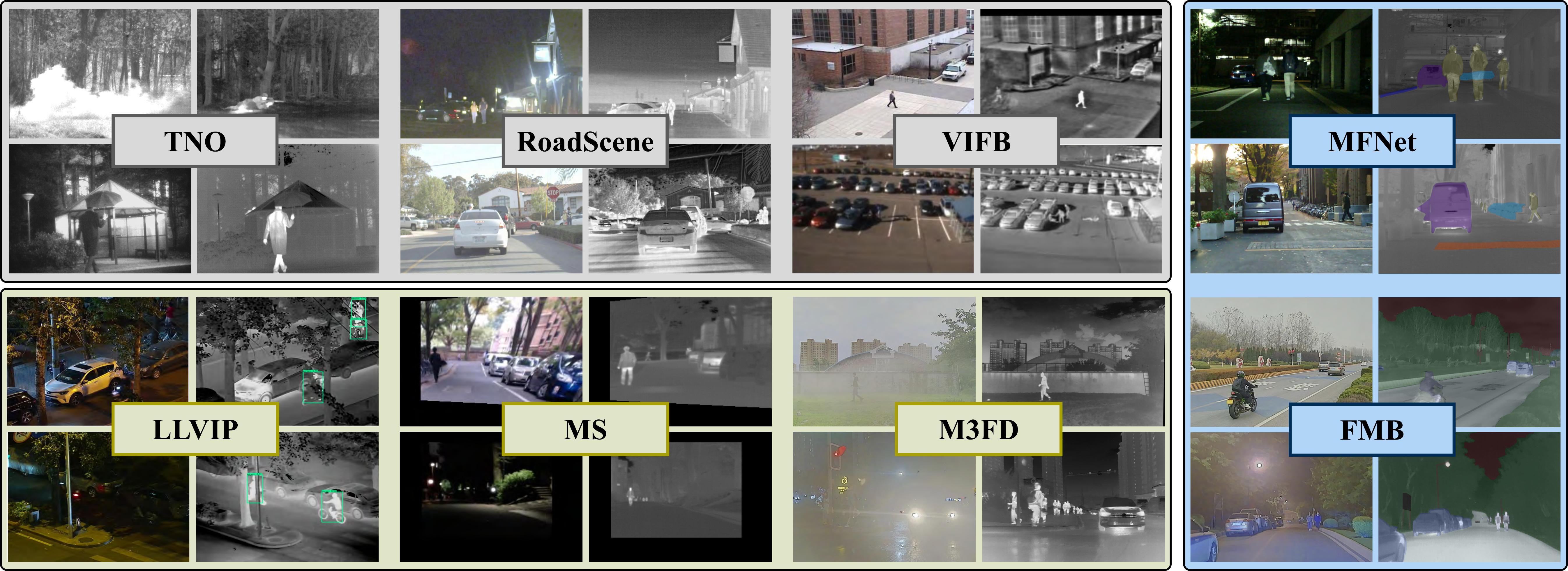}
		\vspace{-0.2cm}
		\caption{An illustration of existing IVIF datasets.}
		\label{fig:data}
		\vspace{-0.5cm}
	\end{figure}
	
		\begin{table*}[!hbt]
		\caption{{Illustration of existing aligned multi-modality datasets.}$^1$}\label{tab:datasettable}
		\vspace{-0.6cm}
		\begin{center}
			\centering
			\renewcommand\arraystretch{0.9} 
			\setlength{\tabcolsep}{2.6mm}
			\begin{tabular}{|llcccccccc|}
				\hline
				&Dataset&Img pairs&Resolution&Color&Camera angle&Nighttime&Objects/Categories&Challenge Scenes&Annotation\\
				\hline
				\rowcolor{gray!10}&TNO~\cite{toet2017tno}&261&768$\times$576& \ding{55}&horizontal&65&few&\ding{52}&\ding{55}\\
				\rowcolor{gray!10}&RoadScene~\cite{xu2020u2fusion}&221&Various&\ding{52}&driving&122&medium&\ding{55}&\ding{55}\\
				\rowcolor{gray!10}&VIFB~\cite{zhang2020vifb}&21&Various&Various&multiplication&10&few&\ding{55}&\ding{55}\\
				\rowcolor{yellow!10}&MS~\cite{takumi2017multispectral}&2999&768$\times$576&\ding{52}&driving&1139&14146~/~6&\ding{55}&\ding{52}\\
				\rowcolor{yellow!10}&LLVIP~\cite{jia2021llvip}&16836&1280$\times$720&\ding{52}&surveillance&all&pedestrian~/~1&\ding{55}&\ding{52}\\
				\rowcolor{yellow!10}&M$^3$FD~\cite{liu2022target}&4200&1024$\times$768&\ding{52}&multiplication&1671&33603~/~6&\ding{52}&\ding{52}\\
				\rowcolor{blue!10}&MFNet~\cite{ha2017mfnet}&1569&640$\times$480&\ding{52}&driving&749&abundant~/~8&\ding{55}&\ding{52}\\
				\rowcolor{blue!10}&FMB~\cite{liu2023multi}&1500&800$\times$600&\ding{52}&multiplication&826&abundant~/~14&\ding{55}&\ding{52}\\
				\bottomrule
			\end{tabular}
		\end{center} 
		\vspace{-0.2cm}
		\footnotesize
		 \begin{flushleft} 
			$^1$ {All dataset download links are organized in the GitHub repository provided in the abstract.}
		\end{flushleft}
		\vspace{-0.6cm}
	\end{table*}

	\subsection{Architectures Summary and Discussion}

	Deep learning-based methods in image fusion are evolving by adopting increasingly complex modules for better modeling, which are categorized by their network structures:
	
	\textbf{Using Existing Architectures.} Many studies~\cite{jian2020sedrfuse, li2021rfn, li2018densefuse} rely on established CNN structures like cascaded, residual, or densely connected blocks, leveraging their proven strengths in feature extraction and information integration.

	\textbf{Complex Stacked Networks.} Methods in this category~\cite{long2021rxdnfuse, mustafa2020infrared, wang2022swinfuse, zhao2021dndt, wang2022res2fusion, ren2021infrared} intertwine different aforementioned blocks, enhancing deep processing for superior fusion quality, and delving into the complementary aspects of various source images for finer fused results.

	\textbf{Multi-Batch Architectures.} Given the modality diversity in infrared-visible fusion, these architectures~\cite{zhao2020didfuse, liu2021learning, sun2022detfusion} incorporate specialized structures to address the varied inputs, using distinct network modules and parameters to optimize the fusion of disparate modal information effectively.

	\textbf{Recursive Architectures.} Emphasizing iterative information enhancement, this group uses recursive designs~\cite{ReCoNet_22} and diffusion models~\cite{zhao2023ddfm} to progressively refine fusion quality, suitable for handling sequential data by capitalizing on prior information for ongoing image improvement.

	Overall, the research in infrared and visible image fusion is moving towards more complex and refined network structures to meet the complex demands fusion.
	\vspace{-0.4cm}
	
	\subsection{Loss Function Summary and Discussion}
	
	In the field of unsupervised infrared and visible image fusion, the design and selection of loss functions are of paramount importance. These functions can generally be understood and classified across three main dimensions: pixel level, evaluative metrics, and data characteristics. (Note that this subsection does not discuss generative models such as GANs and diffusion models.)
	
	At the pixel level, L1 and MSE loss functions~\cite{li2018densefuse, ReCoNet_22, li2023lrrnet} assess image similarity through direct pixel comparisons. SSIM~\cite{li2018densefuse, ReCoNet_22, vs2022image, liu2021learning, li2021rfn}, a key evaluative metric, extends this by considering image structure and quality, reflecting human visual perception. Loss functions targeting data characteristics, like image gradients~\cite{liu2021searching, xu2020fusiondn}, focus on preserving detailed textures within the fusion process.

	\begin{table*}[ht]
	\caption{Quantitative fusion evaluation of several state-of-the-art methods on three datasets (TNO\&RoadScene /M$^3$FD).}
	\vspace{-0.3cm}
	\renewcommand{\arraystretch}{0.9}
	\setlength{\tabcolsep}{1.1mm}{
		\begin{tabular}{|l|c|c|c|c|c|c|c|c|c|c|c|c|c|c|c|c|c|c|}
			\hline
			\multirow{2}{*}{Method}  & \multicolumn{2}{c}{MI} & \multicolumn{2}{c}{VIF} & \multicolumn{2}{c}{CC}   & \multicolumn{2}{c}{SCD}   & \multicolumn{2}{c}{Q$^{\mathrm{AB/F}}$} & \multicolumn{2}{c}{EN}& \multicolumn{2}{c}{SF} & \multicolumn{2}{c}{SD} & \multicolumn{2}{c|}{AG}  \\ \cline{2-19}
			&T\&R &M$^3$  &T\&R &M$^3$ &T\&R &M$^3$ &T\&R &M$^3$ &T\&R &M$^3$ &T\&R &M$^3$ &T\&R &M$^3$ &T\&R &M$^3$ &T\&R &M$^3$ \\ \hline
			BDLFusion~\cite{liu2023bi}   & 2.944   & 2.957 & 0.590 & 0.639 & 0.612 & 0.567 & 1.445 & 1.578 & 0.405 & 0.450 & 7.037 & 6.685 & 8.783 & 8.409 & 37.88 & 29.34 & 3.635 & 3.000  \\
			CAF~\cite{CAF}   & 2.982  &2.982 & 0.585 & 0.612 & 0.606 & 0.541 & 1.531 & 1.654 & 0.495 & 0.511 & 7.122 & 6.923 & 14.00 & 13.55 & 40.55 & 35.26 & 5.172 & 4.366 \\
			CDDFuse~\cite{zhao2023cddfuse}  & 3.113  &3.909 & 0.657 & 0.793 & 0.607 & 0.535 & 1.717 & 1.646 & 0.490 & 0.615 & 7.406 & 6.904 & 17.65 & 14.73 &\cellcolor{blue!15}{54.26} & 37.21 & 6.334 & 4.861 \\
			CoCoNet~\cite{liu2023coconet}  & 2.579  &2.631 & 0.568 & 0.729 & 0.629 & 0.574 & \cellcolor{red!15}{1.782} & 1.772 & 0.363 & 0.380 & \cellcolor{red!15}{7.735} & \cellcolor{red!15}{7.738} & \cellcolor{blue!15}{20.74} & \cellcolor{blue!15}{24.41} & \cellcolor{red!15}{64.44} & \cellcolor{red!15}{62.43} & \cellcolor{blue!15}{7.490} & \cellcolor{blue!15}{7.906} \\
			DATFuse~\cite{tang2023datfuse}   & \cellcolor{blue!15}{3.675}  &4.131 & 0.620 & 0.644 & 0.587 & 0.494 & 1.188 & 1.286 & 0.486 & 0.493 & 6.675 & 6.402 & 11.19 & 10.46 & 31.02 & 26.31 & 3.963 & 3.440 \\
			DDcGAN~\cite{ma2020ddcgan}  & 2.407  &2.542 & 0.439 & 0.602 & 0.592 & 0.540 & 1.499 & 1.665 & 0.321 & 0.481 & 7.430 & \cellcolor{blue!15}{7.424} & 11.38 & 15.63 & 50.40 & 48.89 & 4.368 & 5.578 \\
			DDFM~\cite{zhao2023ddfm}  & 1.857  &2.911 & 0.138 & 0.640 & 0.498 &\cellcolor{red!15}{0.586} & 1.160 & 1.683 & 0.166 & 0.481 & 7.160 & 6.727 & 10.19 & 9.259 & 41.20 & 30.66 & 4.108 & 3.220 \\
			DeFusion~\cite{liang2022fusion}  & 2.955  &3.164 & 0.540 & 0.556 & 0.596 & 0.483 & 1.326 & 1.277 & 0.362 & 0.404 & 6.834 & 6.483 & 7.987 & 8.142 & 34.19 & 27.79 & 3.134 & 2.841 \\
			DenseFuse~\cite{li2018densefuse} & 2.859   &2.914 & 0.583 &  0.603 & 0.591 & \cellcolor{red!15}{0.586} & 1.472 &  1.506 & 0.383 &  0.375 & 6.752 &  6.427 & 8.198 &  7.594 & 31.14 &  25.13 & 2.743 &  2.654 \\
			DetFusion~\cite{sun2022detfusion}   & 2.463  &2.498 & 0.546 & 0.596 & 0.627 & 0.573 & 1.532 & 1.619 & 0.494 & 0.526 & 7.028 & 6.727 & 12.21 & 11.24 & 37.44 & 30.71 & 4.905 & 4.171\\
			DIDFuse~\cite{zhao2020didfuse}  & 2.926  &3.068 & 0.599 & 0.694 & 0.620 & 0.562 & \cellcolor{blue!15}{1.778} & 1.788 & 0.462 & 0.496 & 7.332 & 7.149 & 13.89 & 14.07 & 52.58 & 46.74 & 5.348 & 4.877 \\
			EMMA~\cite{zhao2024equivariant}   & 3.139  &3.816 & 0.643 & 0.769 & 0.596 & 0.502 & 1.652 & 1.494 & 0.449 & 0.592 & 7.035 & 6.910 & 15.23 & 15.22 & 54.25 & 38.27 & 5.919 & 5.338 \\
			FusionDN~\cite{xu2020fusiondn}   & 2.785  &2.961 & 0.579 & 0.703 & 0.606 & 0.561 & 1.681 & \cellcolor{blue!15}{1.809} & 0.462 & 0.511 & 7.418 & 7.335 & 14.57 & 15.01 & 48.40 & 46.38 & 5.789 & 5.354 \\
			FusionGAN~\cite{ma2019fusiongan}  & 2.694  &2.631 & 0.385 & 0.618 & 0.544 & 0.566 & 1.133 & 1.070 & 0.257 & 0.250 & 6.962 & 6.870 & 8.042 & 2.997 & 37.33 & 35.98 & 3.111 & 2.895 \\
			GANMcC~\cite{ma2020ganmcc}  & 2.762  &2.808 & 0.516 & 0.544 & 0.630 & 0.562 & 1.601 & 1.619 & 0.341 & 0.319 & 7.146 & 6.783 & 8.532 & 7.440 & 41.90 & 32.90 & 3.543 & 2.664 \\
			FILM~\cite{zhaoimage}   & 3.101  &\cellcolor{blue!15}{4.294} &  \cellcolor{blue!15}{0.671} & 0.783 & 0.569 & 0.490 & 1.451 & 1.415 & \cellcolor{red!15}{0.573} & 0.626 & 7.250 & 6.869 & 16.05 & 14.86 & 46.61 & 36.39 & 6.007 & 4.836 \\
			IGNet~\cite{ignet}  & 2.106  &2.115 & 0.580 & 0.610 & 0.579 & 0.526 & 1.468 & 1.663 & 0.489 & 0.431 & 7.235 & 7.033 & 13.89 & 12.92 & 44.81 & 41.21 & 5.767 & 4.696 \\
			IRFS~\cite{wang2023interactively}  & 2.691  &2.849 & 0.574 & 0.642 & \cellcolor{red!15}{0.639}&\cellcolor{blue!15}{0.582} & 1.567 & 1.695 & 0.419 & 0.508 & 6.946 & 6.744 & 9.967 & 10.54 & 35.67 & 31.09 & 3.748 & 3.444 \\
			LRRNet~\cite{li2023lrrnet}  & 2.766  &2.805 & 0.508 & 0.566 & 0.594 & 0.542 & 1.558 & 1.463 & 0.352 & 0.498 & 7.118 & 6.437 & 11.92 & 10.69 & 42.54 & 27.20 & 4.497 & 3.601 \\
			MetaFusion~\cite{zhao2023metafusion}   & 2.160  &2.363 & 0.569 & \cellcolor{blue!15}{0.842} & 0.589 & 0.555 & 1.563 & 1.686 & 0.370 & 0.413 & 7.384 & 7.289 & \cellcolor{red!15}{24.85} & \cellcolor{red!15}{25.31} & 52.19 & 44.24 & \cellcolor{red!15}{9.520} & \cellcolor{red!15}{8.833} \\
			MFEIF~\cite{liu2021learning}  & 3.122  &3.114 & 0.633 & 0.660 & \cellcolor{blue!15}{0.633} & 0.578 & 1.617 & 1.647 & 0.460 & 0.482 & 6.991 & 6.676 & 9.147 & 8.764 & 38.40 & 30.12 & 3.590 & 3.097 \\
			MoE-Fusion~\cite{cao2023multi}   & 3.085  &3.491 & 0.613 & 0.701 & 0.553 & 0.487 & 1.450 & 1.492 & 0.509 & 0.579 & 7.043 & 6.993 & 12.17 & 11.65 & 41.95 & 38.49 & 4.818 & 4.241\\
			MRFS~\cite{zhang2024mrfs}  & 2.927  &3.247 & 0.595 & 0.682 & 0.580 & 0.460 & 1.501 & 1.243 & 0.403 & 0.549 & 7.393 & 6.940 & 10.60 & 12.11 & 52.08 & 39.93 & 4.040 & 3.984 \\
			PAIFusion~\cite{liu2023paif}   & 3.225  &3.542 & 0.613 & 0.610 & 0.567 & 0.493 & 1.372 & 1.623 & 0.472 & 0.403 & 6.962 & 6.979 & 11.19 & 9.243 & 39.38 & 36.83 & 4.479 & 3.256\\ 
			PMGI~\cite{zhang2020rethinking}  & 3.077  &3.117 & 0.483 & 0.495 & 0.582 & 0.522 & 0.784 & 1.229 & 0.216 & 0.255 & 6.266 & 6.305 & 5.440 & 6.183 & 22.03 & 23.47 & 2.207 & 2.185 \\
			PromptF~\cite{liu2024promptfusion}   & 3.418  &4.127 &  \cellcolor{red!15}{0.700} & 0.786 & 0.601 & 0.503 & 1.640 & 1.487 & 0.500 & 0.608 & 7.333 & 6.800 & 15.31 & 13.59 & 51.11 & 34.79 & 5.571 & 4.520\\
			PSFusion~\cite{tang2023rethinking}  & 2.608  &2.741 & 0.631 & 0.826 & 0.611 & 0.559 & 1.722 & \cellcolor{red!15}{1.831} & 0.536 & 0.573 & \cellcolor{blue!15}{7.445} & 7.399 & 16.54 & 20.71 & 51.20 & \cellcolor{blue!15}{49.56} & 6.594 & 6.913 \\
			ReCoNet~\cite{ReCoNet_22}  & 2.985  &3.066 & 0.540 & 0.577 & 0.578 & 0.470 & 1.510 & 1.438 & 0.376 & 0.485 & 7.051 & 6.679 & 10.00 & 13.63 & 41.88 & 35.81 & 3.810 & 4.205 \\
			RFN-Nest~\cite{li2021rfn}  & 2.677  &2.881 & 0.526 & 0.583 & 0.631 & 0.572 & 1.737 & 1.727 & 0.316 & 0.406 & 7.282 & 6.864 & 7.297 & 7.724 & 44.80 & 33.64 & 3.164 & 2.856 \\
			SDNet~\cite{zhang2021sdnet}  & 3.064  &3.219 & 0.569 & 0.554 & 0.564 & 0.501 & 1.211 & 1.420 & 0.494 & 0.517 & 6.935 & 6.662 & 12.21 & 12.10 & 35.64 & 31.26 & 4.855 & 4.202 \\
			SeAFusion~\cite{tang2022image}  & 3.016  &3.574 & 0.619 & 0.722 & 0.596 & 0.524 & 1.594 & 1.586 & 0.497 & 0.598 & 7.299 & 6.846 & 15.82 & 13.95 & 49.01 & 35.44 & 6.211 & 4.780 \\
			SegMiF~\cite{liu2023multi}   & 2.811  &3.052 & 0.669 & 0.804 & 0.611 & 0.559 & 1.683 & 1.744 & 0.540 & 0.653 & 7.257 & 6.983 & 14.78 & 14.25 & 49.18 & 37.70 & 5.684 & 4.822 \\
			SHIP~\cite{zheng2024probing}   & \cellcolor{red!15}{3.782} & \cellcolor{red!15}{4.812} & 0.661 & 0.819 & 0.556 & 0.466 & 1.324 & 1.310 & \cellcolor{blue!15}{0.548} & 0.643 & 7.127 & 6.824 & 15.55 & 15.25 & 42.81 & 35.23 & 5.868 & 5.175\\
			SuperFusion~\cite{SuperFusion_22}   & 3.451 &3.445 & 0.622 & 0.627 & 0.580 & 0.533 & 1.457 & 1.545 & 0.487 & 0.479 & 7.092 & 6.757 & 12.06 & 10.31 & 42.68 & 32.64 & 4.365 & 3.498\\
			SwinFusion~\cite{ma2022swinfusion}   & 3.345 & 4.161 & 0.645 & 0.774 & 0.602 & 0.521 & 1.599 & 1.561 & 0.468 & 0.611 & 6.974 & 6.803 & 11.89 & 13.65 & 43.47 & 35.77 & 4.438 & 4.602\\
			TarDAL~\cite{liu2022target}  & 3.241 & 3.161 & 0.569 & 0.602 & 0.575 & 0.510 & 1.443 & 1.551 & 0.427 & 0.406 & 7.287 & 7.150 & 13.30 & 12.56 & 46.78 & 43.02 & 4.653 & 4.144 \\
			Text-IF~\cite{yi2024text}   & 2.930 & 3.685 & 0.658 &  \cellcolor{red!15}{0.896} & 0.586 & 0.499 & 1.581 & 1.518 & 0.536 & \cellcolor{red!15}{0.675} & 7.352 & 6.931 & 15.83 & 15.62 & 49.62 & 36.66 & 6.250 & 5.271\\
			TGFuse~\cite{rao2023tgfuse}  & 2.718 & 3.503 & 0.645 & 0.833 & 0.567 & 0.463 & 1.430 & 1.333 & 0.535 & \cellcolor{blue!15}{0.658} & 7.149 & 6.802 & 13.64 & 14.59 & 43.38 & 35.43 & 5.032 & 4.888 \\
			TIMFusion~\cite{liu2023task}   & 3.656 & 2.745 & 0.635 & 0.543 & 0.553 & 0.310 & 1.281 & 0.554 & 0.402 & 0.498 & 7.081 & 6.712 & 11.93 & 13.42 & 41.37 & 39.01 & 4.361 & 4.348\\
			U2Fusion~\cite{xu2020u2fusion}  & 2.599 & 2.760 & 0.556 & 0.633 & 0.621 & 0.567 & 1.338 & 1.569 & 0.489 & 0.539 & 6.821 & 6.659 & 11.10 & 10.71 & 32.11 & 28.83 & 4.545 & 3.966 \\
			UMFusion~\cite{UMF}  & 2.888 & 3.089 & 0.610 & 0.613 & 0.625 & 0.546 & 1.475 & 1.570 & 0.470 & 0.398 & 6.967 & 6.699 & 10.17 & 8.758 & 36.06 & 30.53 & 3.875 & 2.928 \\
			YDTR~\cite{tang2022ydtr}  & 2.976 & 3.183 & 0.588 & 0.635 & 0.620 & 0.554 & 1.420 & 1.506 & 0.436 & 0.478 & 6.842 & 6.547 & 10.03 & 10.10 & 34.84 & 28.00 & 3.684 & 3.304 \\
			\hline
		\end{tabular}
		}
	\label{tab:T_NUM1}
	\vspace{-0.5cm}
\end{table*}

	Building on these foundations, more complex variants of loss functions have emerged to address specific challenges in image fusion. For instance, the application of Visual Salience Map (VSM~\cite{liu2023multi, liu2023bi, liu2022target}) at the pixel level represents an innovation, resulting in more nuanced fusion effects. At the evaluative metrics level, using less common metrics like Spatial Frequency (SF)~\cite{tang2022ydtr} as loss functions emphasizes the frequency characteristics and visual effects of images, thus achieving effective fusion while maintaining visual comfort. Further, complex image feature-based loss functions, such as the maximization of gradients or edge extraction~\cite{UMF, ma2022swinfusion, tang2023datfuse, ignet, tang2023rethinking} and including perceptual and contrastive losses~\cite{liu2023coconet}, provide deeper insights and solutions.
	
	The design of loss functions is diverse, and apart from the above-mentioned types, many specially designed losses have been introduced. Researchers can select, combine, and optimize these based on the characteristics of source images and task requirements, driving further development and innovation in the field of image fusion.
	\vspace{-0.2cm}

	\section{Benchmark and Evaluation Metric}
	\vspace{-0.1cm}
	\subsection{Benchmark}
	With the advancement in the field of infrared and visible image fusion, numerous datasets have been proposed and utilized. They can be broadly categorized into three types: early fusion-oriented datasets, fusion datasets aimed at target detection, and fusion datasets for semantic segmentation. Table~\ref{tab:datasettable} provides a detailed overview of their respective characteristics, while Figure~\ref{fig:data} displays typical image pairs from them. 
	
			\vspace{-0.2cm}

		\begin{figure*}[t]
			
		\centering
		\setlength{\tabcolsep}{1pt} 
		
		\includegraphics[width=0.98\textwidth]{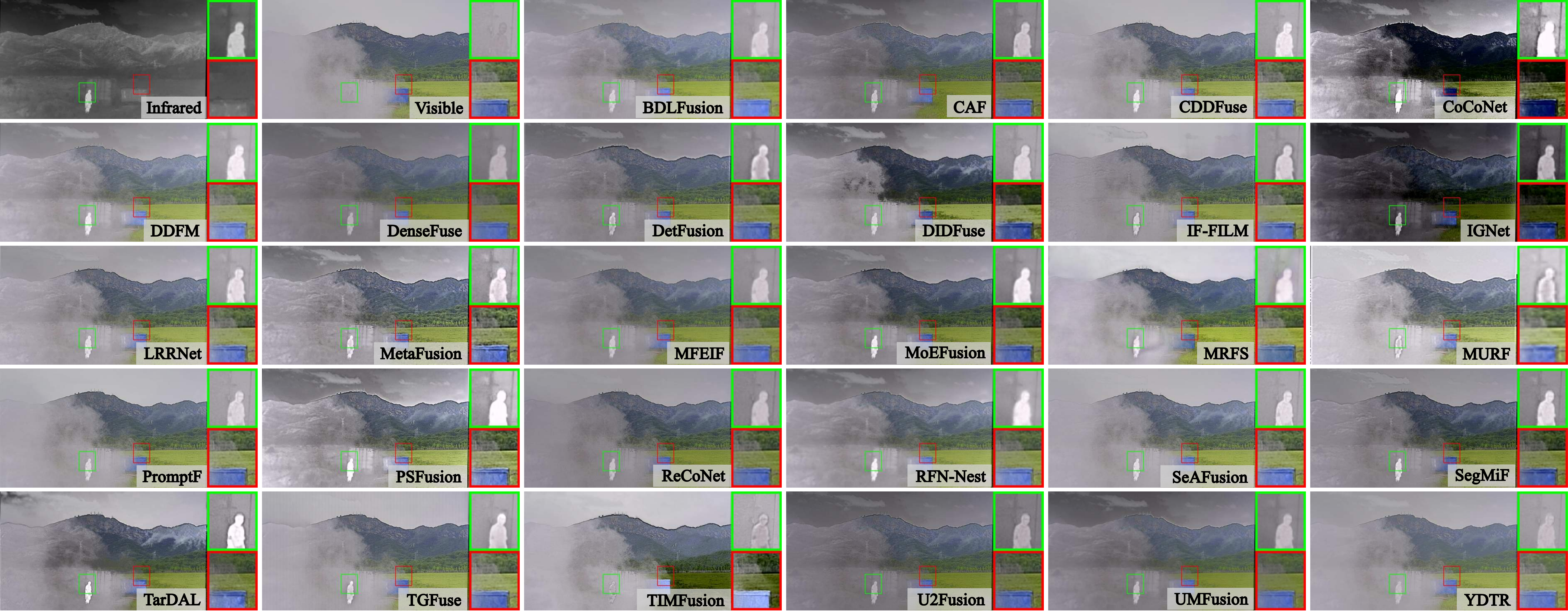}
		\vspace{-0.2cm}
		\caption{{Compared with several state-of-the-art fusion methods on a typical image pair of M$^3$FD.}}
		\label{fig:fusion}
		\vspace{-0.4cm}
	\end{figure*}

\vspace{-0.1cm}
	
	\subsection{Evaluation Metric}
	\label{set: 1}
	\subsubsection{Fusion-oriented Metric}
	This section summarizes 9 fusion metrics: 5 reference-based (MI, VIF, CC, SCD, Q$^\text{AB/F}$) and 4 no-reference metrics (EN, SF, SD, AG).
	
	Mutual Information \textbf{(MI)}~\cite{Qu2002Information} measures the information transferred from source to fused images. 
	Visual Information Fidelity \textbf{(VIF)}~\cite{han2013new} assesses fusion fidelity aligned with the human visual system, with higher values indicating better performance.
	Correlation Coefficient \textbf{(CC)}~\cite{ma2019infrared}  evaluates how a fused image mirrors its sources, focusing on linear correlation. In contrast, Sum of Difference Correlation \textbf{(SCD)} measures the integration of unique information from source images. CC emphasizes existing relationships, while SCD targets new elements.
	Gradient-Based Fusion Performance \textbf{(Q$^\text{AB/F}$)}~\cite{xydeas2000objective} assesses edge detail preservation, with scores near 1 indicating effective edge retention.
	Entropy \textbf{(EN)}~\cite{Roberts2008Assessment} gauges information content in fused images but is sensitive to noise. 
	Spatial Frequency \textbf{(SF)}~\cite{eskicioglu1995image} evaluates detail and texture sharpness, indicating richer edge and texture information with higher values.
	Standard Deviation \textbf{(SD)}~\cite{Rao1997In-fibre} reflects image quality in terms of distribution and contrast, where higher contrast leads to greater SD values. 
	Average Gradient \textbf{(AG)}~\cite{Cui2015Detail} measures texture features and details, with higher AG values indicating enhanced fusion performance.

	\subsubsection{Registration Metric}
	The registered results are subjected to evaluation utilizing three widely accepted metrics, namely Mean Squared Error \textbf{(MSE)}~\cite{wang2009mean}, Mutual Information \textbf{(MI)}, and Normalized Cross Correlation \textbf{(NCC)}~\cite{NCC_metric}.
	
	~\textbf{MSE} measures the average squared difference between pixels of two images, assessing their alignment. A lower MSE value signifies greater similarity, making it a crucial metric in image registration and fusion.~\textbf{MI} introduced above is also a prevalent similarity metric within image registration. A higher MI means that the two images are well-aligned.~\textbf{NCC} is a metric that assesses the similarity between corresponding windows within two images for evaluating registration accuracy.  
	\subsubsection{Perception Metric}
	This part introduces key metrics for segmentation and detection tasks.

	In semantic segmentation, Intersection over Union \textbf{(IoU)} measures the overlap between predicted and true areas, with mean IoU \textbf{(mIoU)} averaging across categories. Accuracy \textbf{(Acc)} indicates the proportion of correctly classified pixels, while mean Accuracy \textbf{(mAcc)} averages this across categories.

	In object detection, Recall and Precision, based on IoU thresholds, define true positives. Average Precision \textbf{(AP)} measures single-class performance, and mean Average Precision \textbf{(mAP)} averages AP across all classes.

	\vspace{-0.2cm}
	\section{Performance Summary}
	
	\subsection{Image Fusion}
	
	{In this section, we employ the three most commonly used fusion datasets (TNO, RoadScene, and M$^3$FD) to compare the performance of various advanced fusion methods, using the pre-trained models released by the original authors.}
	\subsubsection{Qualitative Comparisons}
	
	{We selected typical challenging scenario images from M$^3$FD to evaluate the visualization effects of fusion, as shown in Figure~\ref{fig:fusion}. A major difficulty in this scenario is smoke, which is a challenging yet unnecessary element. IGNet avoids most of its impact but thus leans excessively towards the infrared image. While CoCoNet highlights information on people and background vehicles, it exhibits color deviation in the smoke-free background (the grassland in the red box). In contrast, CAF and MoEFusion partially avoid the impact of smoke while preventing color deviation.}
	
	\subsubsection{Quantitative Comparisons}
	{The 9 fusion metrics introduced in Section~\ref{set: 1} are used to compare quantitative results, where 57 pairs from TNO, 221 pairs from RoadScene, and 300 pairs from the M$^3$FD fusion dataset are randomly selected for calculation. The specific metrics are shown in Table~\ref{tab:T_NUM1}.}
	
	{In supervised metrics, methods using traditional source
	image-based loss functions have a significant advantage (e.g.,
	CC of DenseFuse) due to their thorough preservation of source image information. Methods optimized for segmentation also show advantages in visual fidelity metrics like VIF (e.g., PromptF), as they fully leverage semantic information. In unsupervised metrics, CoCoNet, utilizing feature-level contrastive loss to optimize
	the entire training process, achieves representation capabilities far beyond conventional loss functions, thereby standing out in performance.}

	\subsection{Registration-based Image Fusion}
	We survey 7 misaligned multi-modality image fusion methods including UMFusion~\cite{UMF}, SuperFusion~\cite{SuperFusion_22}, ReCoNet~\cite{ReCoNet_22}, RFVIF~\cite{RFVIF_23}, SemLA~\cite{SemLA_23}, MURF~\cite{MURF_23}, and IMF~\cite{IMF_23}, and evaluate their fusion performance with slight deformation using 9 widely-used metrics.

	\begin{figure*}[!htb]
		\centering
		\setlength{\tabcolsep}{1pt} 
		
		\includegraphics[width=0.98\textwidth]{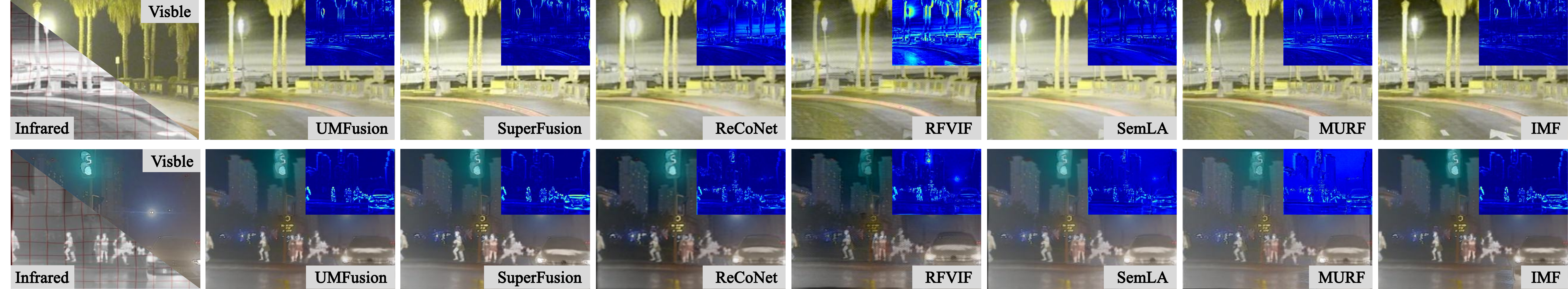}
		\vspace{-0.2cm}
		\caption{{Misaligned \cellcolor compared with several state-of-the-art fusion methods on two typical image pairs.}}
		\label{fig:reg}
		\vspace{-0.5cm}
	\end{figure*}
	
	\begin{table*}[t]
		\begin{center}
			\caption{Quantitative fusion evaluation of cross-modality misaligned images on the Roadscene and M$^3$FD datasets.}
			\vspace{-0.4cm}
			\label{tab:reg-fusion}
			\renewcommand{\arraystretch}{0.9}
			\resizebox{1.0\linewidth}{!}{
				\begin{tabular}{|l|ccccccccc|ccccccccc|}
					\hline
					\multirow{2}{4em}{{Methods}} &\multicolumn{9}{c}{{RoadScene}}&\multicolumn{9}{c|}{{M$^3$FD}} \\ \cline{2-19}
					& MI & VIF & CC & SCD & Q$^{\mathrm{AB/F}}$ & EN & SF & SD & AG & MI & VIF & CC & SCD & Q$^{\mathrm{AB/F}}$ & EN & SF & SD & AG\\
					\hline
					UMFusion~\cite{UMF} & 1.91 & 0.61 & \cellcolor{red!15}{0.60} & \cellcolor{blue!15}{1.45} & 0.25 & 6.92 & 11.75 & 34.09 & 6.46 & \cellcolor{blue!15}{2.17} & \cellcolor{red!15}{0.87} & \cellcolor{red!15}{0.61}& 1.29 & 0.22 & 6.58 & 9.67 & 27.68 & 5.23 \\
					SuperFusion~\cite{SuperFusion_22} & \cellcolor{red!15}{2.25} & \cellcolor{blue!15}{0.62} & 0.54 & 1.39 & \cellcolor{red!15}{0.29} & 7.08 & 14.43 & \cellcolor{blue!15}{41.62} & 7.47 & 2.09 & 0.71 & \cellcolor{red!15}{0.61} & \cellcolor{blue!15}{1.37} & \cellcolor{red!15}{0.33} & 6.61 & 12.74 & 29.95 & 6.44 \\
					ReCoNet~\cite{ReCoNet_22} & \cellcolor{blue!15}{2.12} & 0.53 & 0.55 & 1.44 & 0.24 & 7.02 & 10.71 & 39.98 & 6.34 & 1.95 & 0.41 & 0.58 & 1.26 & 0.24 & 6.60 & 7.75 & 30.45 & 4.72 \\
					RFVIF~\cite{RFVIF_23} & 1.55 & 0.14 & 0.46 & 1.19 & 0.07 & \cellcolor{red!15}{7.37} &\cellcolor{blue!15}{15.92} &\cellcolor{red!15}{46.05} & \cellcolor{red!15}{8.42} & 1.63 & 0.16 & 0.54 & 1.25 & 0.12 & \cellcolor{red!15}{7.16} & 12.54 & \cellcolor{red!15}{40.71} & \cellcolor{blue!15}{7.24} \\
					SemLA~\cite{SemLA_23} & 1.81 & 0.53 & 0.52 & 1.26 & 0.15 & 6.89 & 14.86 & 36.55 & 5.44 & 1.94 & 0.48 & 0.56 & 1.23 & 0.16 & 6.74 & \cellcolor{blue!15}{14.08} & 31.99 & 4.41 \\
					MURF~\cite{MURF_23} & 1.83 & 0.42 & 0.56 & 1.40 & 0.23 & \cellcolor{blue!15}{7.17} & \cellcolor{red!15}{16.00} & 40.79 & \cellcolor{blue!15}{8.00} & 1.88 & 0.31 & 0.57 & 1.28 & 0.22 & \cellcolor{blue!15}{6.78} & 12.84 & \cellcolor{blue!15}{32.32} & 6.21 \\
					IMF~\cite{IMF_23} & 2.10 & \cellcolor{red!15}{0.64} & \cellcolor{blue!15}{0.59} & \cellcolor{red!15}{1.51} & \cellcolor{blue!15}{0.28} & 7.10 & 13.74 & 38.94 & 6.78 & \cellcolor{red!15}{2.27} & \cellcolor{blue!15}{0.79} & 0.54 & \cellcolor{red!15}{1.54} & \cellcolor{blue!15}{0.31} & 6.69 & \cellcolor{red!15}{21.07} & 30.51 & \cellcolor{red!15}{7.88} \\
					\hline
			\end{tabular}}
		\end{center}
		\vspace{-0.4cm}
	\end{table*}

	\subsubsection{Qualitative Comparisons}
	Figure~\ref{fig:reg} shows qualitative results of various fusion methods confronted with misaligned infrared and visible images.
	
	{By observation, the style transfer-based registration and fusion methods~(UMFusion and IMF) and the supervised SuperFusion achieved comparable results, effectively correcting structural distortion and edge ghosts. In contrast, the latent space-based methods~(RFVIF, SemLA and ReCoNet) still exhibited some residual deformations in the fused images.}
	
	{Additionally, the results suggest that IMF, MURF, and UMFusion exhibit superior rankings among the top three methods in terms of object saliency and texture richness in the fused images.}
	
	\subsubsection{Quantitative Comparisons}
	Table~\ref{tab:reg-fusion} reports quantitative of misaligned infrared and visible images on the RoadScene and M$^3$FD datasets.
	
	Due to the use of ground truth deformation fields as supervision, SuperFusion exhibits excellent performance on several reference-based metrics such as MI, VIF, CC, and Q$^{\mathrm{AB/F}}$.
	
	Moreover, style transfer-based registration-then-fusion methods (\textit{e.g.}, UMFusion and IMF) achieve suboptimal results on reference-based metrics.
	
	In contrast, latent feature space-based methods outperform in several no-reference metrics (\textit{e.g.}, EN, SF, SD, and AG).
	
	This phenomenon suggests that style transfer-based and latent feature space-based methods have their own strengths. The development of a framework capable of harnessing their respective advantages holds tremendous potential.
	
	\vspace{-0.2cm}
	\subsection{Image Fusion for Object Detection}
	{In this part, we employ YOLO-v5 as the base detector, using fusion images from all methods in a unified training setup, with M$^3$FD as the dataset for both training and testing.}
	
	Due to the presence of a large number of continuously captured images in M$^3$FD, a simple random sampling method for dividing the dataset can easily lead to overfitting. 
	In this study, we randomly selected 100 sets of 10 consecutive images as the test set, where the scenes are almost non-existent in the training set, to validate the robustness of the fusion method.
	
	\begin{figure*}[!htb]
		\centering
		\setlength{\tabcolsep}{1pt} 
		
		\includegraphics[width=0.98\textwidth]{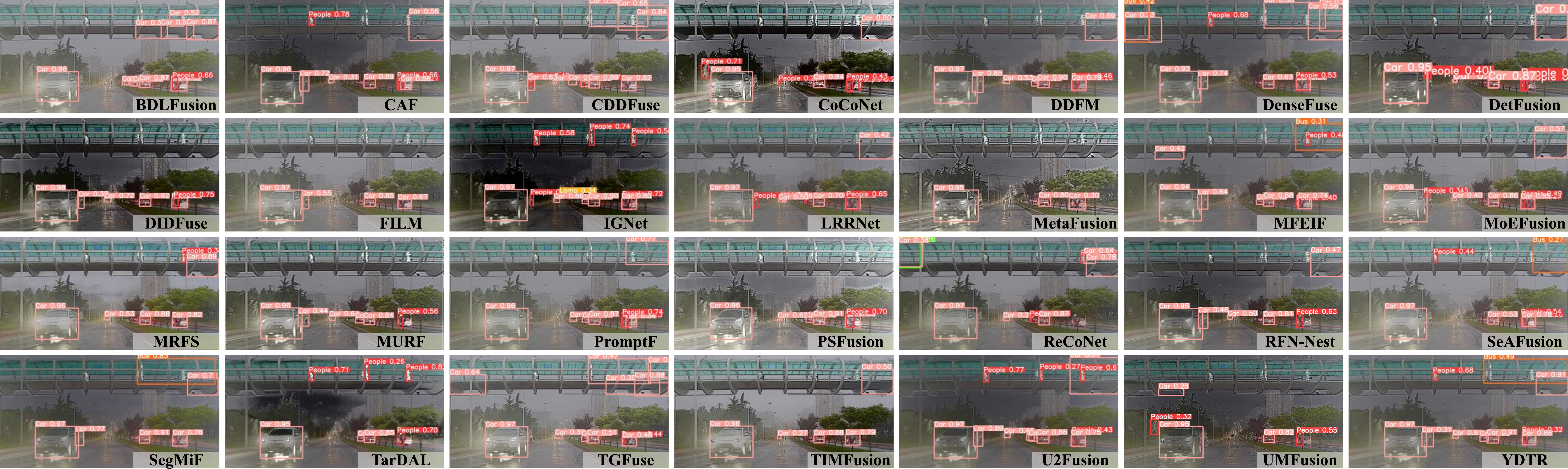}
		\vspace{-0.2cm}
		\caption{{Object detection results based on image fusion compared with several state-of-the-art  methods.}}
		\label{fig:result_ir_vis_od}
		\vspace{-0.5cm}
	\end{figure*}

	\subsubsection{Qualitative Comparisons} As shown in {Figure~\ref{fig:result_ir_vis_od} shows the detection precision of different fusion methods in a challenging small-target scenario. Two main factors affect detection: the scene suffers from degradation (fog, rain, strong light) and weak textures make pedestrians on the bridge difficult to detect. Methods like DDFM, LRRNet and ReCoNet produce low-contrast fused images, failing to highlight infrared targets, while methods like DDcGAN and IRFS generate artifacts that lead to incorrect detection. However, TarDAL and PAIFusion effectively detect all pedestrians. This highlights the importance of complementary information (e.g., thermal targets) and handling degradation for improved detection performance.}
	
		\begin{table}[htb]
		\centering
		\footnotesize
		
		\renewcommand{\arraystretch}{0.9}
		\setlength{\tabcolsep}{1.1mm}{
			\vspace{-0.4cm}
			\caption{{Quantitative  results of object detection on the M$^3$FD  dataset.}}~\label{tab:detec}
			\vspace{-0.3cm}
			\begin{tabular}{|l|ccccccc|}
				\hline
				{Methods}
				& Lamp & Car & Bus & Motor & Truck & People & mAP \\ \hline
				BDLFusion~\cite{liu2023bi} & 0.340 & 0.809 & 0.695 & 0.358 & 0.392 & 0.666    &  0.544    \\ 
				CAF~\cite{CAF} & 0.391 & 0.827 & 0.702 &0.367 & 0.460  & 0.713    &  0.577     \\
				CDDFuse~\cite{zhao2023cddfuse} & 0.427 & 0.821 & 0.704 & 0.329 & 0.453 & 0.665     &   0.566  \\ 
				CoCoNet~\cite{liu2023coconet} & 0.360 & 0.827  & 0.719 & 0.308 & 0.468   & 0.708 &       0.469   \\
				DDcGAN~\cite{ma2020ddcgan} & 0.407 &  0.801 & 0.666 &  0.300 &0.392 & 0.603   &    0.528   \\ 
				DDFM~\cite{zhao2023ddfm} & 0.462 & 0.831 & 0.712 & 0.339 & 0.469 & 0.686  &  0.583 \\ 
				DenseFuse~\cite{li2018densefuse} &  0.464 &0.820 &  0.710 &  0.310&  0.483 &  0.668   &     0.576   \\ 
				DetFusion~\cite{sun2022detfusion} &0.419&0.830&0.700&0.403&0.447&0.679&0.580  \\ 
				DIDFuse~\cite{zhao2020didfuse} & 0.406 & 0.813 & 0.713 & 0.311 & 0.444 & 0.681    &   0.561  \\ 
				EMMA~\cite{zhao2024equivariant} & 0.462 & 0.827 & 0.647 &0.392 & 0.473  & 0.682    &  0.581     \\
				FusionGAN~\cite{ma2019fusiongan} & 0.399  & 0.708 &0.708 & 0.301 & \cellcolor{blue!15}{0.486}  & 0.693 &     0.569     \\ 
				FILM~\cite{zhaoimage} & 0.439 & 0.821 & 0.691 &0.354 & \cellcolor{red!15}{0.492}  & 0.674    &  0.578     \\
				IGNet~\cite{ignet} & 0.360 & 0.827  & 0.719 & 0.308 & 0.468   & 0.708    &  0.565  \\ 
				IRFS~\cite{wang2023interactively} & 0.370 &0.822  &0.687  &   0.278   &   0.444   & 0.615   &   0.536  \\ 
				LRRNet~\cite{li2023lrrnet} & 0.461 & 0.828 & 0.696 & 0.331 & 0.454 & 0.664     &   0.572 \\ 
				MFEIF~\cite{liu2021learning} &0.403 & 0.810 &\cellcolor{blue!15}{0.741} & 0.296& 0.467 & 0.563 &   0.563   \\ 
				{MoEFusion}~\cite{cao2023multi} &  \cellcolor{blue!15}{0.469} & \cellcolor{blue!15}{0.837}    & 0.690 &    0.333  &  0.469    &0.676    &   0.579 \\
				MRFS~\cite{zhang2024mrfs} & 0.420 & 0.812 & 0.688 &\cellcolor{blue!15}{0.424} & 0.471  & 0.655    &  0.578     \\
				MURF~\cite{MURF_23} & 0.416 & 0.788 & 0.698& 0.239 & 0.425 & 0.643   &  0.535    \\ 
				PAIFusion~\cite{liu2023paif} & 0.327 & 0.811 & 0.722 & 0.219 & 0.440& \cellcolor{red!15}{0.728} &   0.541     \\ 
				PromptF~\cite{liu2024promptfusion} & 0.414 & 0.818 & 0.638 &0.378 & 0.454  & 0.656    &  0.560     \\
				PSFusion~\cite{tang2023rethinking} & 0.415 & 0.828 & 0.688 & 0.334 & 0.464& 0.691 &      0.570   \\ 
				ReCoNet~\cite{ReCoNet_22} & 0.409 & 0.797 & 0.689 & 0.299 & 0.477 & 0.644     & 0.552  \\ 
				RFN-Nest~\cite{li2021rfn} & \cellcolor{red!15}{0.497} & 0.830 & 0.702 & 0.385&0.438& 0.671 &   \cellcolor{blue!15}{0.587}  \\  
				SDNet~\cite{zhang2021sdnet} & 0.377 & \cellcolor{red!15}{0.846} & \cellcolor{red!15}{0.743} & 0.356 & 0.485 & 0.715    &  \cellcolor{blue!15}{0.587}   \\ 
				SeAFusion~\cite{tang2022image} & 0.398 & 0.826 & 0.719 & 0.340 & 0.471 & 0.694    &   0.575  \\ 
				SegMiF~\cite{liu2023multi} & 0.468 & 0.831 & 0.727 & 0.365 & 0.438 & 0.684      &   0.585   \\ 
				SHIP~\cite{zheng2024probing} & 0.427 & 0.832 & 0.666 &0.419 & 0.480  & 0.687    &  0.585     \\
				SuperFusion~\cite{SuperFusion_22} & 0.393 &0.823 & 0.706&0.336 &0.473  &0.696    &  0.571      \\ 
				SwinFusion~\cite{ma2022swinfusion} & 0.409 & 0.826 & 0.717 & 0.308& 0.516 & 0.686 &    0.577 \\ 
				TarDAL~\cite{liu2022target} & 0.451 & 0.824 & 0.740 & 0.369 & 0.477 & \cellcolor{blue!15}{0.727}      & \cellcolor{red!15}{0.598}  \\ 
				Text-IF~\cite{yi2024text} & 0.398 & 0.814 & 0.681 &0.394 & 0.453  & 0.654    &  0.566     \\
				TGFuse~\cite{rao2023tgfuse} & 0.242 &0.787 & 0.541 &0.209 & 0.412 & 0.611     & 0.467       \\ 
				TIMFusion~\cite{liu2023task} &  0.431 & 0.823 & 0.699 & \cellcolor{red!15}{0.479} & 0.453 & 0.622 &    0.584   \\ 
				U2Fusion~\cite{xu2020u2fusion} & 0.433 & 0.835 & 0.710 & 0.348 & 0.461 & 0.704 & 0.582  \\ 
				UMFusion~\cite{UMF} & 0.411  & 0.830 & 0.713 & 0.281  & 0.457  & 0.717 &     0.568   \\ 
				YDTR~\cite{tang2022ydtr} &  0.338  &  0.820 &  0.682  & 0.263    &   0.415    &  0.604  &      0.520   \\   \hline
			\end{tabular}
		}
		
	\end{table}

	\subsubsection{Quantitative Comparisons} 
	{We present the detection results in Table~\ref{tab:detec}, with AP@0.5 for different classes and overall precision, recall, mAP@0.5, and mAP@0.5:0.95. Two key observations emerge: First, TarDAL, a detection-oriented fusion method, delivers the highest precision by preserving thermal target details and textural information, excelling in detecting cars and persons. Second, perception-guided methods like SegMiF and TIMFusion show strong performance through semantic feature interjection and task loss guidance, while U2Fusion and SDNet balance information preservation across modalities. }

	\subsection{Image Fusion for Semantic Segmentation}
	
	{We uniformly use the advanced SegFormer~\cite{xie2021segformer} as the base segmenter to measure the performance of various advanced fusion methods, with all methods trained on the corresponding fused images under consistent settings.} The official division of training and testing sets in FMB is applied.
	
	\subsubsection{Qualitative Comparisons}
\begin{figure*}[t]
		\centering
		\setlength{\tabcolsep}{1pt} 
		
		\includegraphics[width=0.96\textwidth]{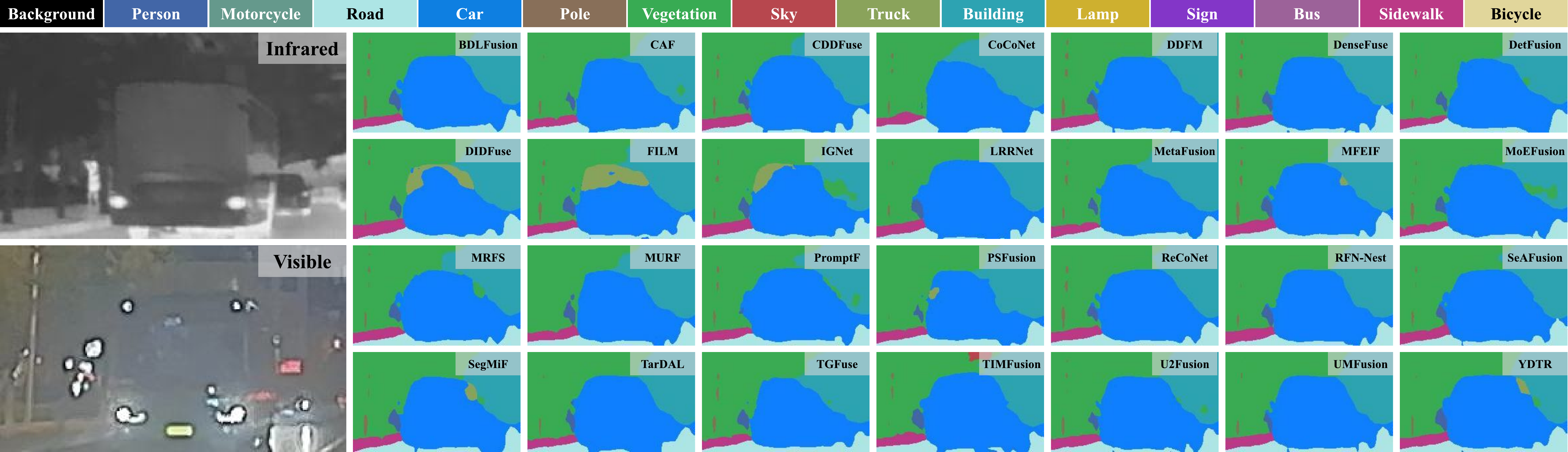}
		\vspace{-0.2cm}
		\caption{{Semantic segmentation results based on image fusion compared with several state-of-the-art  methods.}}
		\label{fig:seg}
		\vspace{-0.4cm}
	\end{figure*}

	{Figure~\ref{fig:seg} presents a qualitative comparison of a typical night-time urban scene. In this scenario, a pedestrian located in a dark area is challenging for most methods to segment accurately due to the low visibility. Meanwhile, the bus emits such strong light that it causes overexposure in the scene, leading many methods to misclassify the bus as a car and fail to capture its full outline. These challenges highlight the ongoing difficulty of handling scenes with extreme lighting conditions.
	}

	\subsubsection{Quantitative Comparisons}
	In Table~\ref{tab:seg}, we report the numerical results for the segmentation task. {Similar to the trend observed in detection tasks, perception-guided fusion methods (e.g., MoEFusion and DetFusion) demonstrate competitive segmentation performance due to their task-specific loss incorporation or learning strategies.}

		\begin{table*}[h]
		\vspace{-0.1cm}
		\centering
		\footnotesize
		\caption{{Quantitative semantic segmentation results of different methods on the {FMB} dataset.}}~\label{tab:seg}
		\vspace{-0.5cm}
		\renewcommand{\arraystretch}{0.9}
		\setlength{\tabcolsep}{1.7mm}{
			\begin{tabular}{|l|cc|cc|cc|cc|cc|cc|cc|cc|cc|}
				\hline
				\multirow{2}{*}{Methods}& \multicolumn{2}{c}{Person}  & \multicolumn{2}{c}{Car} &			\multicolumn{2}{c}{Road}&  \multicolumn{2}{c}{Sidewalk} & \multicolumn{2}{c}{Building} & \multicolumn{2}{c}{Sign}&\multicolumn{2}{c}{Bus}&
				\multicolumn{2}{c|}{Motocycle}
				&\multirow{2}{*}{mAcc}&\multirow{2}{*}{mIoU} \\ \cline{2-17} 
				&Acc &IoU & Acc &IoU& Acc &IoU& Acc &IoU& Acc &IoU& Acc &IoU& Acc &IoU& Acc &IoU& &\multicolumn{1}{c|}{} \\
				\hline
				BDLFusion~\cite{liu2023bi} & 76,6 & 64.7 & 93.4 & 81.1 & 96.0 & 88.6 & 57.6 & 48.8 & 92.0 &\cellcolor{blue!15}{83.7} & 84.8 & 69.4 & 68.7 & 63.5 &39.5 & 24.6 &64.2 & 56.0 \\
				CAF~\cite{CAF} & 76.8 & 64.7 & 93.3 & 81.0 & 96.1 & 88.6 & 56.3 & 48.5 & 91.3 & 83.3 & 85.9 & 71.2 & 59.4 & 53.8 & 38.5 & 26.2 & 63.2 & 55.5 \\
				CDDFuse~\cite{zhao2023cddfuse} & 76.4 & 63.4 & 93.9 & 81.6 & 96.0 & 88.0 & 54.5 & 45.7 & 90.8 &81.8 & 84.6 & 70.9 & 61.0 & 56.6 & 48.4 & 26.1 & 64.3 & 55.4 \\
				CoCoNet~\cite{liu2023coconet} & 72.5 & 60.5 & \cellcolor{red!15}{94.3} & 81.7 & 95.9 & 88.4 & 53.0 & 46.0 & 92.0 & 83.1 & 84.2 & 70.4 & 67.3 & 63.2 & 42.7 & 26.7 & 64.2 & 56.4 \\
				DATFuse~\cite{tang2023datfuse} & 74.8 & 62.7 & 93.5 & 80.2 & 95.9 & 87.7 & 51.1 & 43.8 & 90.8 & 81.7 & 83.8 &68.8 & 57.4 &52.9 & 37.9 & 21.7 & 61.9 & 53.9 \\
				DDcGAN~\cite{ma2020ddcgan} & 74.9 & 61.8 & 93.1 & 80.8 & 95.7 & 87.5 & 50.8 & 41.7 & 90.9 & 81.9 & 83.1 & 70.7 & 59.6 & 57.2 & 33.5 & 26.9 & 62.2 & 54.9 \\
				DDFM~\cite{zhao2023ddfm} & 74.1 & 63.3 & 93.5 & 81.6 & \cellcolor{blue!15}{96.4} & \cellcolor{red!15}{89.1} & 54.9 & 48.3 & \cellcolor{blue!15}{92.4} & \cellcolor{red!15}{84.0} & 85.5 & 71.7 & 68.1 & 63.3 & 48.0 & 25.7 & \cellcolor{red!15}{64.9} & \cellcolor{blue!15}{56.6} \\
				DeFusion~\cite{liang2022fusion} & 76.7 &63.3 & 93.4 & 81.2 & \cellcolor{blue!15}{96.4} & 87.5 & 51.1 & 44.6 & 90.7 & 82.3 & 85.3 & 70.7 & \cellcolor{blue!15}{71.5} & 66.0 & \cellcolor{blue!15}{50.4} & 27.5 & 64.5 & 55.9 \\
				DenseFuse~\cite{li2018densefuse} & 75.3 & 63.6 & 93.4 & 81.8 & \cellcolor{red!15}{96.5} & 88.6 & 52.2 & 45.6 & 92.1 &\cellcolor{blue!15}{83.7} & 85.9 & 71.3 & \cellcolor{red!15}{71.8} & \cellcolor{red!15}{67.0} & 45.0 & 24.0 & \cellcolor{blue!15}{64.8} & 56.5 \\
				DetFusion~\cite{sun2022detfusion} & 76.3 & 64.4 & 93.6 & 81.1 & 95.8 & 88.7 & 58.6 & \cellcolor{blue!15}{49.2} & 91.7 & 83.4 & 84.3 & 70.2 & 60.4 & 56.8 & 41.1 & 27.0 & 64.6 & 56.5 \\
				DIDFuse~\cite{zhao2020didfuse} & 75.4 & 64.0 & 93.8 & 79.6 & 95.7 & 88.9 & \cellcolor{red!15}{59.9} & 46.5 & 91.2 & 82.4 & 84.6 & 70.3 & 51.2 & 48.2 & 43.1 & 25.2 & 62.3 & 54.0 \\
				EMMA~\cite{zhao2024equivariant} & 75.9 & 63.1 & 93.8 & 81.0 & 95.9 & 88.3 & 49.4 & 42.9 & 91.5 & 82.8 & 84.7 & 71.1 & 66.8 & 62.9 & 46.8 & 27.0 & 64.2 & 56.0 \\
				FusionDN~\cite{xu2020fusiondn} & 75.6 & 63.2 &\cellcolor{blue!15}{94.1} & 81.3 & 95.9 & 88.6 & 55.3 & 47.9 & 91.9 & 83.3 & 85.6 & 71.2 & 66.5 & 61.3 & 45.2 & 26.3 & 63.7 & 55.7 \\
				FusionGAN~\cite{ma2019fusiongan} &\cellcolor{blue!15}{77.9} & 64.7 & 93.1 & 79.1 & 95.5 & 87.5 & 54.7 & 46.6 & 91.4 & 82.3 & 85.6 & \cellcolor{red!15}{72.4} & 42.9 & 39.2 & 35.0 & 23.3 & 62.0 & 54.3 \\
				GANMcC~\cite{ma2020ganmcc} & 73.8 & 63.7 & 92.9 & 80.9 & 96.0 & 88.1 & 57.8 & \cellcolor{blue!15}{49.2} & 91.5 & 82.3 & 84.4 & 70.4 & 53.8 & 50.2 & 46.7 & 28.9 & 63.4 & 55.3 \\
				FILM~\cite{zhaoimage} & 75.9 & 64.4 & 93.7 & 80.4 & 96.1 & 88.7 & 51.9 & 44.5 & 91.9 & 83.3 & 84.3 & 71.8 & 62.4 & 58.1 & 43.8 & 23.8 & 63.6 & 55.5 \\
				IGNet~\cite{ignet} & 74.1 & 63.3 & 93.7 & 81.0 & 95.9 &\cellcolor{blue!15}{89.0} & \cellcolor{blue!15}{59.0}& \cellcolor{red!15}{50.1} & 92.0 &83.5 &86.2 &70.7 & 69.2 & 62.2 &46.0 & 22.4 & 63.8 & 55.3 \\
				IRFS~\cite{wang2023interactively} & 74.2 & 62.2 & 93.6 & 81.2 & 96.0 & 88.8 & 56.3 & 44.4 & 91.0 & 82.5 & 85.8 & 72.0 & 68.8 & 63.8 & 47.1 & 25.5 & \cellcolor{blue!15}{64.8} & 56.1 \\
				LRRNet~\cite{li2023lrrnet} & 74.9 & 63.6 & 93.7 & 81.6 & 96.1 & 88.9 & 55.2 & 47.0 & 91.8 & 82.6 & 85.6 & 71.1 & 68.4 & 64.3 & 46.0 & 27.5 & 64.5 & 56.4 \\
				MetaFusion~\cite{zhao2023metafusion} & 75.8 & 63.4 & 93.9 & 81.3 & 96.0 & 88.8 & 54.9 & 45.0 & 90.8 & 82.4 & 85.8 & 71.0 & 63.5 & 59.6 & 44.8 & 27.0 & 63.7 & 55.7 \\
				MFEIF~\cite{liu2021learning} & 75.8 & 64.4 & 93.5 & 81.2 &96.2 & 88.6 & 55.5 & 47.5 & 92.0 & 83.1 & 86.3 & 70.4 & 67.6 & 62.7 & 48.1 & 26.3 & 64.7 & 56.1 \\
				MoEFusion~\cite{cao2023multi} & 75.2 & 63.8 & 93.5 & \cellcolor{blue!15}{81.9} & 96.1 & 88.7 & 55.8 & 46.7 & \cellcolor{red!15}{94.2} & 83.1 & 85.4 & \cellcolor{blue!15}{72.2} & 70.8 & \cellcolor{blue!15}{66.4} & 40.1 & 30.7 & 64.5 & \cellcolor{red!15}{57.0} \\
				MRFS~\cite{zhang2024mrfs} & 76.2 & 63.8 & 93.1 & 81.1 & 96.3 & 88.2 & 49.4 & 43.6 & 92.0 & 82.7 & 85.7 & 70.8 & 61.6 & 56.8 & 43.1 & 27.3 & 62.4 & 54.6 \\
				MURF~\cite{MURF_23} & 72.2 &61.5 & 93.5 & 79.5 & 96.3 & 87.7 & 48.6 & 43.1 & 91.3 & 81.7 & 85.4 & 70.9 & 57.8 & 54.2 & 41.9 & 24.6 & 61.7 & 54.0 \\
				PAIFusion~\cite{liu2023paif} & 75.8 & 64.7 & 93.6 & 81.4 & \cellcolor{blue!15}{96.4} & 88.8 & 55.5 & 47.9 & 92.0 & 83.3 & \cellcolor{red!15}{87.1} & 71.9 & 65.6 & 59.8 & 45.4 & 24.4 & 64.2 & 55.9 \\
				PMGI~\cite{zhang2020rethinking} & \cellcolor{red!15}{78.2} & \cellcolor{red!15}{66.0} &93.6 & \cellcolor{red!15}{82.5} & 95.6 & 88.7 & 58.2 & 48.0 & 90.7 & 81.5 & 76.7 & 65.1 & 68.8 & 64.7 & 43.8 & \cellcolor{red!15}{31.9} & 63.4 & 56.1 \\
				PromptF~\cite{liu2024promptfusion} & 77.2 & 63.8 & 93.6 & 80.7 & 96.2 & 88.2 & 52.7 & 45.8 & 91.2 & 82.2 & 84.0 & 70.4 & 59.3 & 55.5 & 46.2 & 27.9 & 63.5 & 55.4 \\
				PSFusion~\cite{tang2023rethinking} & 76.6 & 64.2 & 93.5 & 80.7 & 96.0 & 88.2 & 52.5 & 45.5 & 91.6 & 83.0 & 83.9 & 72.0 & 52.1 & 49.1 & 49.0 & 29.2 & 63.8 & 55.7 \\
				ReCoNet~\cite{ReCoNet_22} & 74.3 & 62.3 & 93.8 & 81.2 & 95.7 & 88.8 & 57.8 & 48.9 & 91.6 & 82.6 & 85.5 & 71.6 & 68.9 & 64.0 & 46.4 & 27.7 & 64.5 & 56.4 \\
				RFN-Nest~\cite{li2021rfn} & 76.2 & 63.2 & 93.2 & 80.8 & 96.2 & 88.5 & 54.5 & 46.0 & 92.2 & 83.6 & \cellcolor{blue!15}{86.6} & \cellcolor{red!15}{72.4} & 60.8 & 56.7 & 48.6 & 26.3 &64.7 & 56.0 \\
				SDNet~\cite{zhang2021sdnet} & 75.6 & 64.3 & 93.5 & 81.5 & 96.2 & 88.3 & 54.8 & 47.6 & 92.2 & 83.2 & 85.7 & 71.6 & 63.9 & 58.6 & 43.6 & 26.1 & 64.2 & 56.2 \\
				SeAFusion~\cite{tang2022image} & 76.8 & 64.7 & 93.8 & 81.1 & 96.2 & 88.4 & 55.1 & 46.9 & 91.7 & 82.6 & 83.5 & 70.5 & 69.1 & 63.9 & 44.9 & 23.9 & 63.1 & 55.0 \\
				SegMiF~\cite{liu2023multi} & 75.3 & 64.2 & 93.4 & 80.9 & 95.9 & 88.4 & 54.5 & 47.6 & 91.6 & 82.8 & 85.1 & 71.0 & 59.5 & 55.4 & 45.2 & 28.7 & 64.0 & 56.0 \\
				SHIP~\cite{zheng2024probing} & 76.1 & 64.5 & 93.8 & 80.9 & 96.0 & 88.2 & 52.3 & 43.9 & 91.9 & 83.0 & 85.1 & 71.7 & 59.9 & 56.8 & 44.9 & 25.3 & 63.9 & 55.7 \\
				SuperFusion~\cite{SuperFusion_22} & 77.3 & 64.0 & 93.6 & 80.6 & 96.1 & 87.7 & 53.2 & 44.5 & 91.0 & 82.6 & 84.9 & 71.3 & 67.0 & 61.7 & 41.7 & 26.3 & 64.0 & 55.9 \\
				SwinFusion~\cite{ma2022swinfusion} & 77.3 & 64.1 & 93.6 & 81.3 & 96.0 & 88.7 & 54.6 & 47.4 & 91.6 & 82.3 & 83.9 & 70.8 & 68.5 & 63.7 & 43.8 & 26.0 & 64.7 & 56.4 \\
				TarDAL~\cite{liu2022target} & 76.0 & \cellcolor{blue!15}{64.8} & 93.2 & 79.7 & 96.0 & 87.8 & 52.8 & 45.0 & 91.9 & 82.9 & 86.4 & 71.4 & 56.4 & 53.0 & 43.7 & 26.9 & 62.1 & 54.2 \\
				Text-IF~\cite{yi2024text} & 77.1 & 64.6 & 93.9 & 80.8 & 96.1 & 88.7 & 54.9 & 47.2 & 91.7 & 83.0 & 85.6 & 71.9 & 64.6 & 60.6 & 45.1 & 28.5 & 64.6 & 56.4 \\
				TGFuse~\cite{rao2023tgfuse} & 77.7 & 64.7 & 93.5 & 80.7 & 96.2 & 87.9 & 51.9 & 44.6 & 90.8 & 82.5 & 85.7 & 72.1 & 64.3 & 60.1 & \cellcolor{red!15}{51.2} &\cellcolor{blue!15}{31.6} & 64.4 & 56.1 \\
				TIMFusion~\cite{liu2023task} & 71.9 & 62.1 & 92.9 & 81.3 & 95.5 & 88.0 & 53.4 & 45.5 & 90.6 & 81.4 & 82.0 & 69.9 & 59.0 & 55.5 & 41.7 & 25.5 & 62.7 & 55.0 \\
				U2Fusion~\cite{xu2020u2fusion} & 76.3 & 64.0 & 93.7 & 81.3 & 96.1 & 88.8 & 55.9 & 46.1 & 91.4 & 83.3 & 85.7 & 71.9 & 61.4 & 57.7 & 43.5 & 26.0 & 64.1 & 56.0 \\
				UMFusion~\cite{UMF} & 75.8 & 64.4 & 93.7 & 81.8 & 95.9 & 88.9 & 58.7 & 47.2 & 91.7 & 82.9 & 86.3 & 71.3 & 67.9 & 64.2 & 41.7 & 24.2 & 64.6 & 56.3 \\
				YDTR~\cite{tang2022ydtr} & 74.3 & 63.4 & 93.7 & 81.7 & 95.9 & 88.9 & 58.7 & 49.1 & 91.9 & 83.2 & 85.3 & 70.6 & 64.6 & 60.2 & 44.2 & 28.0 & 64.5 & 56.5 \\
				\hline
				
		\end{tabular} }
		\vspace{-0.2cm}
	\end{table*}

	\subsection{Computational Complexity Analysis}
	
	In this part, we select three key metrics for comparing computational efficiency: average runtime, the number of network model parameters, and FLOPS (floating-point operations per second). Average runtime focuses on the time required for an algorithm or model to complete a specific task, serving as an intuitive measure of speed and efficiency. FLOPS, on the other hand, is a measure of the computational volume of a model, indicating the number of floating-point operations during forward propagation. The parameters, which includes the total count of weights and biases, reflects the size and complexity of the model. Models with more parameters may exhibit stronger learning capabilities but could also lead to higher computational and storage costs. Generally, the parameter count of a model tends to be proportional to its FLOPS, and an increase in FLOPS usually results in longer runtime.
	
	For our experiments, we tested random sets of 10 images from the M$^3$FD dataset, each with a resolution of 1024$\times$768, on an Nvidia GeForce 4090. To eliminate the influence of the CPU, we utilized the official event function from CUDA to measure the runtime on the GPU, excluding the initial value to calculate the average. The final results in units of \emph{ms, M, and G} are presented in Table~\ref{tab:time}. It is evident that there are significant differences in average runtime, FLOPS, and the number of parameters among various methods. For instance, Densefuse, an extremely simple and outdated method, manages to achieve a very low runtime while maintaining a lower count of parameters and FLOPS. In contrast, DDFM, which utilizes a diffusion model, far exceeds other methods in terms of the number of parameters and FLOPS, and its runtime is as high as 280.8k ms. This may indicate that the efficiency of this method in practical applications is relatively low.

		\begin{table*}[]
		\centering
		\vspace{-0.2cm}
		\caption{{Quantitative comparison of computational efficiency on M$^3$FD dataset.}}
		\vspace{-0.3cm}
		\label{tab:time}
		\renewcommand{\arraystretch}{0.9}
		\setlength{\tabcolsep}{1.0mm}{
			\begin{tabular}{|llll|llll|llll|}
				\toprule[1.0pt]
				{Method}    & {Time\tiny{(ms)}} & {FLOPS\tiny(G)} & {Params\tiny(M)} & {Method}    & {Time\tiny(ms)} & {FLOPS\tiny(G)} & {Params\tiny(M)} & {Method}    & {Time\tiny(ms)} & {FLOPS\tiny(G)} & {Params\tiny(M)} \\
				\midrule[1.0pt]
				BDLFusion~\cite{liu2023bi} & 45.89    & 328.2    & 0.418   & IGNet~\cite{ignet}     & 47.46    & 647.6    & 7.871    & SeAFusion~\cite{tang2022image} & 23.53    & 130.5    & 0.167   \\
				CAF~\cite{CAF} & 11.38    & 116.7    & 0.148   &IRFS~\cite{wang2023interactively}      & 18.21    & 188.8    & 0.242         & SegMiF~\cite{liu2023multi}     & 681.5    & 500.4    & 0.621    \\
				CDDFuse~\cite{zhao2023cddfuse}   & 463.8    & 1402     & 1.186  & LRRNet~\cite{li2023lrrnet}     & 263.9    & 36.27    & 0.049     & SHIP~\cite{zheng2024probing}     & 27.93    & 401.5    & 0.525      \\
				CoCoNet~\cite{liu2023coconet}     & 29.06    & 498.5    & 9.114 & MetaFusion~\cite{zhao2023metafusion}     & 85.40    & 637.9    & 0.812  & SuperFusion~\cite{SuperFusion_22}     & 177.0    & 196.3    & 1.962   \\
				DATFuse~\cite{tang2023datfuse}     & 33.06    & 14.22    & 0.011   &MFEIF~\cite{liu2021learning}     & 68.93    & 554.7    & 0.371     & SwinFusion~\cite{ma2022swinfusion}     & 62.74    & 185.1    & 0.955    \\
				DDFM~\cite{zhao2023ddfm}      & 280.8k   & 1336k    & 552.7   & MoE-Fusion~\cite{cao2023multi} & 77.79    & 229.5    & 83.45   & TarDAL~\cite{liu2022target}     & 16.57    & 233.3    & 0.297    \\
				DeFusion~\cite{liang2022fusion}  & 62.04    & 196.8    & 7.874  & MRFS~\cite{zhang2024mrfs}     & 98.90    & 356.0    & 135.0      & Text-IF~\cite{yi2024text}     & 21.37    & 135.7    & 85.49  \\
				DetFusion~\cite{sun2022detfusion}  & 64.15    & 187.0    & 83.07  & PAIFusion~\cite{liu2023paif} & 87.72    & 204.8    & 0.260        & TGFuse~\cite{rao2023tgfuse}    & 41.17    & 318.9    & 137.3     \\
				DIDFuse~\cite{zhao2020didfuse}  & 12.23    & 45.61    & 2.733  & PSFusion~\cite{tang2023rethinking}  & 79.66    & 463.6    & 45.90     & TIMFusion~\cite{liu2023task}     & 17.39    & 100.7    & 0.158   \\
				EMMA~\cite{zhao2024equivariant}  & 25.72    & 106.3    & 1.516    &ReCoNet~\cite{ReCoNet_22}   & 37.53    & 18.21    & 0.008       &UMFusion~\cite{UMF}  & 41.21    & 494.6    & 0.629         \\
				FILM~\cite{zhaoimage}  & 183.3    & 230.4    & 0.490   &RFN-Nest~\cite{li2021rfn}  & 178.8    & 1346     & 7.524 & YDTR~\cite{tang2022ydtr}      & 143.3    & 247.0    & 0.107      \\
				
				\bottomrule[1.0pt]
			\end{tabular}
		}
		\vspace{-0.6cm}
	\end{table*}
	
	\section{Future Trends}
	\subsection{Handling of Misalignment/Attack Data}
	
	Image registration is a key factor for fusion, ensuring that corresponding pixel points in different images are accurately aligned. In 
	practical application, due to the different imaging principles (reflection and radiation) and spectral ranges, obtaining pair of images with aligned pixels is challenging. However, few efforts have been focus on this issues~\cite{UMF, SuperFusion_22, ReCoNet_22, RFVIF_23, SemLA_23, MURF_23, IMF_23}. We anticipate that future endeavors should focus on designing fusion networks that are compatible with registration or incorporating registration as a component of the overall loss function, thereby leading to the creation of ``robust" fusion network.
	
	The robustness of image fusion networks in complex adversarial scenarios, including both physical (involving distortions and degradation) and digital attacks (such as parameter perturbations), represents a significant challenge. The core issue revolves around current methodologies that focus mainly on preserving source details without fully addressing the correlation between fusion elements and perceptual robustness. Additionally, there is a noticeable deficiency in methods that enhance robustness against adversarial conditions without compromising overall performance, with prevalent issues like loss fluctuations and pattern failures highlighting the need for network and training optimization. To improve security, integrating visible and infrared sensors has become common to counteract single-modal physical attacks. While PAIFusion~\cite{liu2023paif} has explored parameter perturbation in image fusion, effectively countering real-world physical and especially cross-modality attacks~\cite{wei2023unified,wei2023physically} remains a formidable task.

	\subsection{Developing Benchmarks}
	
	High-quality benchmarks are vital for the IVIF community's development, initially highlighted by sets like TNO, RoadSenece, and VIFB, which had limited scene diversity and resolution. To advance perception tasks, datasets such as MS and LLVIP were developed, offering object detection labels but with a focus on specific scenes like roads and surveillance. Addressing broader challenges, M$^3$FD introduce with adverse weather conditions and varied scenarios. Recent datasets like MFNet and FMB cater to complex semantic tasks with extensive labels. While the former encompasses 8 categories specific to driving scenes, the latter boasts 14 categories spanning multiple environments.
	
	Although the aforementioned benchmarks have alleviated data scarcity in the IVIF domain, three pressing issues still warrant attention. i) The creation of infrared and visible image registration benchmarks is crucial, as existing benchmarks focus on pixel-aligned pairs. Developing benchmarks that accurately reflect real-world discrepancies with registration ground truth, considering factors like imaging differences and baselines, is essential for discipline progression. ii) Broadening IVIF benchmarks to include a variety of high-level tasks are key to advancing the field. While current benchmarks cover tasks like object detection and semantic segmentation, emerging needs call for the inclusion of tasks (such as depth estimation and scene parsing) to meet the evolving research directions and practical applications. iii) Exploring diverse challenging scenarios: Existing benchmarks predominantly focus on limited scenes, with the majority of data sourced from urban locales and academic settings. This encourages us to explore more demanding situations, such as tunnels and caves (for exploration), forest terrains (for military surveillance), and indoor spaces (for rescue missions), among others.
	
	\subsection{Better Evaluation Metrics}
	
	Evaluating the quality of fused images is a critical issue, especially in the absence of ground truth. Traditional metrics like EN, MI, CC, and SCD each measure only one aspect of image quality and may not align with subjective evaluations, particularly under conditions like high noise levels. As a result, these metrics alone cannot fully capture the essence of image fusion quality. 
	
	While leveraging subsequent task performance, such as object detection or semantic segmentation, provides insights, it is difficult to comprehensively reflect visual quality. Therefore, developing evaluation metrics that account for both visual and perceptual quality is an essential research direction.

	\subsection{Lightweight Design}
	Operating efficiency is a pivotal factor for the practical application of IVIF. Current methodologies strive to achieve high-quality fused imagery by increasing neural network size, which significantly boosts the number of parameters and adversely affects operational efficiency. While some strategies have been proposed to mitigate these issues by integrating innovative technologies, such as Neural Architecture Search, Neural Network Pruning (NNP), and Atrous Convolution (AC), their effectiveness is still heavily dependent on the computational power of advanced GPUs.

	In light of these challenges, future research should place greater emphasis on the pursuit of lightweight network design. This involves addressing two critical issues: firstly, the construction of lightweight networks is imperative since most existing devices (e.g., unmanned aerial vehicles (UAVs) and handheld devices) are unable to support the computational demands of heavy GPUs. This necessitates the exploration of more efficient and less resource-intensive network architectures. Moreover, the development of hardware-friendly fusion methods is essential, as they present a more economical and practical solution compared to heavy image computing units. The preponderance of extant methods, which have been developed on server-based platforms utilizing frameworks such as PyTorch and TensorFlow, are not easily transferable to actual devices/products. Therefore, ensuring that new methods are adaptable to actual hardware constraints is vital for real-world application and integration.
	
	\subsection{Combination with Various Tasks}
	IVIF as a basic image enhancement technology, undoubtedly can boost/assist/combine with other vision tasks. i)~Advanced Scene Analysis: in autonomous vehicle navigation, combining these two types of images can provide a more comprehensive understanding of the environment, aiding in better decision-making and obstacle avoidance in varied lighting and weather conditions. ii)~Depth Estimation and 3D Reconstruction: IVIF can also enhance depth estimation and 3D reconstruction tasks. With combination of infrared/visible complementary depth information, it can lead to more accurate 3D models of environments. This has potential applications in virtual reality, augmented reality, and urban planning.  As computational technologies continue to evolve, the potential of IVIF in transforming various fields is immense, promising to unlock new capabilities and applications in the near future. 
	
	\section{Conclusions}
	In the realm of infrared and visible image fusion, there have been notable advancements.  However, uncertainties remain regarding technology alignment with various data types, practical applications, and evaluation standards. In this survey, we clarify these aspects from methodological and practical perspectives, and include a detailed comparative analysis of registration, fusion, and related tasks. The goal of this survey is to provide guidance for both beginners and seasoned professionals in this evolving domain, encouraging the continued development and application of these technologies. We also spotlight areas ripe for future research, with the hope of inspiring ongoing innovation in this dynamic field.

	\bibliographystyle{IEEEtran}
	\bibliography{reference}
	\vspace{-1.5cm}
	\begin{IEEEbiography}[{\includegraphics[width=1in,height=1.25in,clip,keepaspectratio]{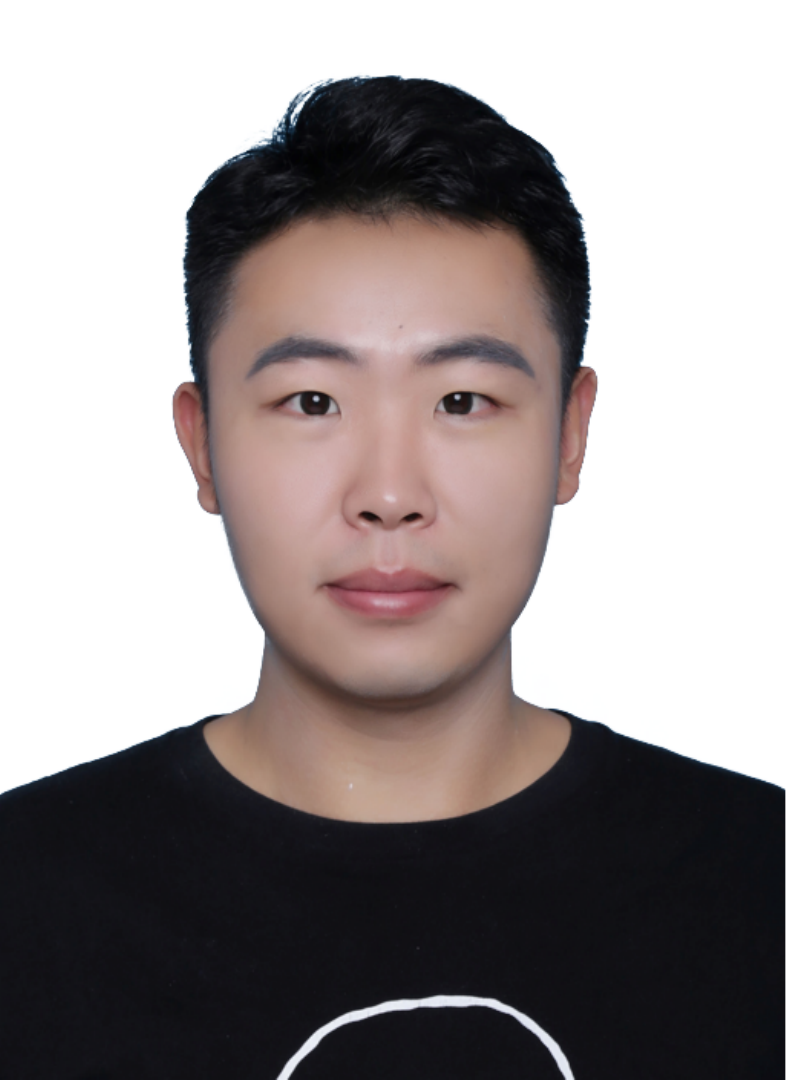}}]{Jinyuan Liu} obtained his M.S. degree in Computer Science from Dalian University, Dalian, China, in 2018. He completed his Ph.D. degree in Software Engineering at the Dalian University of Technology, Dalian, in 2022. He is currently a Postdoctoral Fellow in the School of Mechanical Engineering at Dalian University of Technology. His research interests include computer vision, image fusion, and deep learning.
	\end{IEEEbiography}
	\vspace{-1cm}
	\begin{IEEEbiography}[{\includegraphics[width=1in,height=1.25in,clip,keepaspectratio]{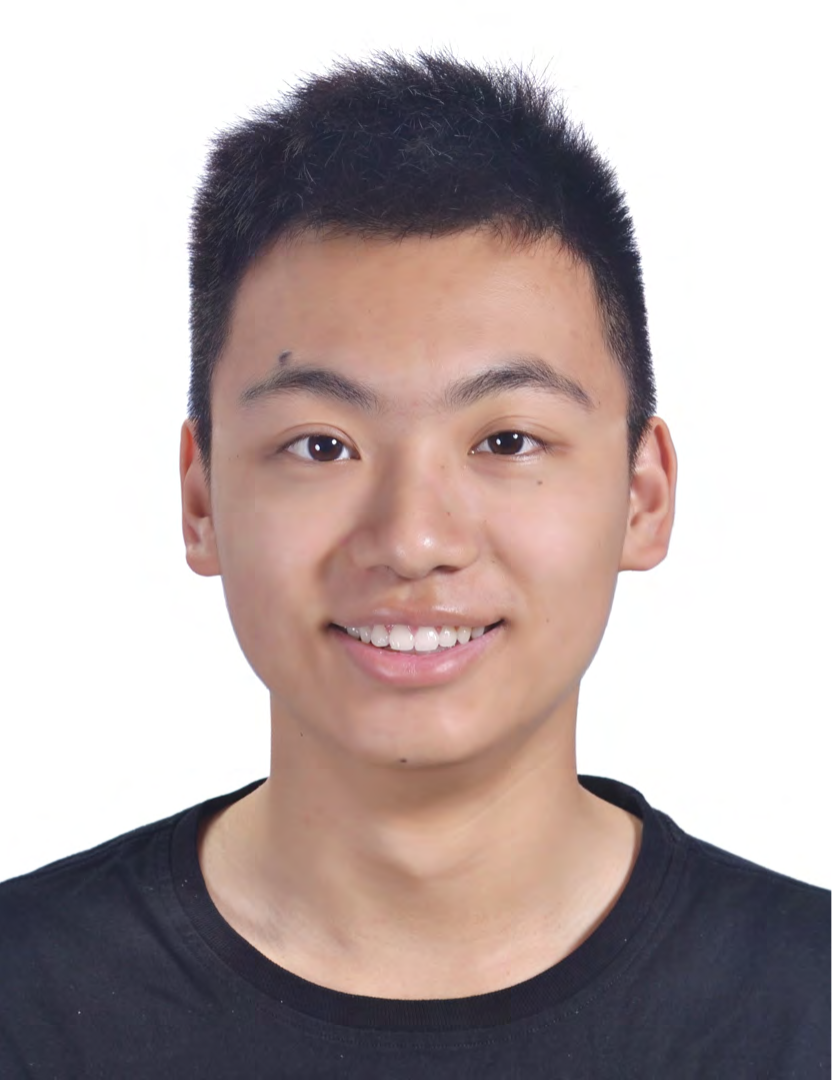}}]{Guanyao Wu} received his B.S. degree from the DUT-RU International School of Information Science and Engineering, Dalian University of Technology, China, in 2022. He is currently enrolled in an integrated master's and Ph.D. program in Software Engineering at the same institution. His research interests include deep learning-based computer vision, with a focus on image fusion and processing.
		
	\end{IEEEbiography}
	\vspace{-1cm}
	\begin{IEEEbiography}[{\includegraphics[width=1in,height=1.25in,clip,keepaspectratio]{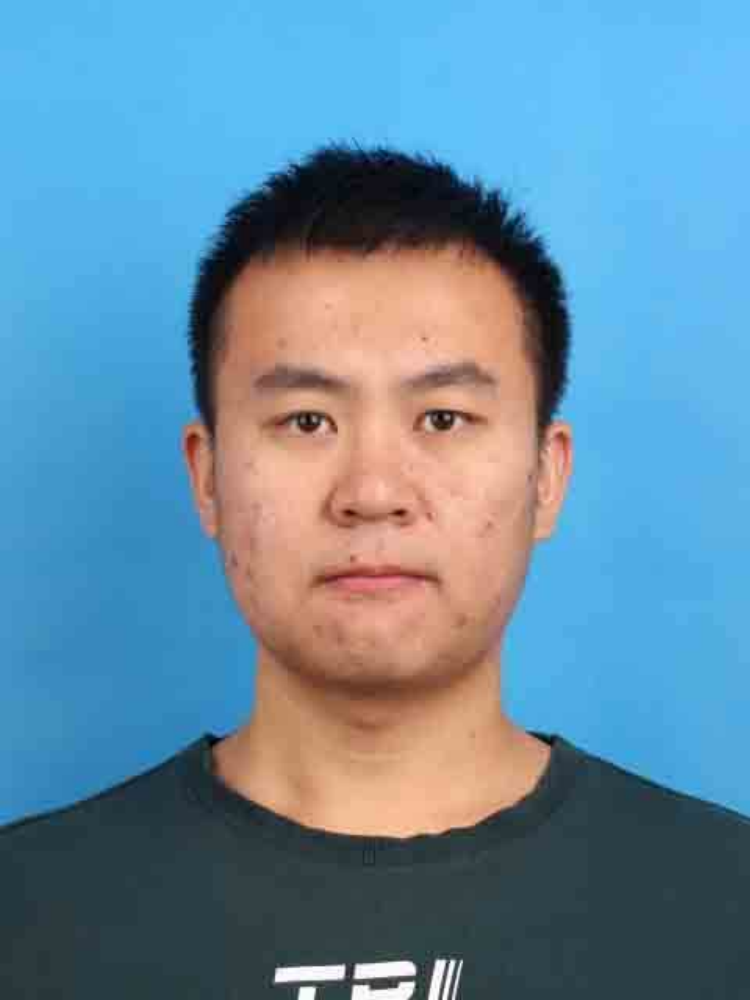}}] {Zhu Liu} received the B.S. degree in software
	engineering from the Dalian University of Technology, Dalian, China, in 2019. He received his M.S. degree in Software Engineering at Dalian University of Technology, Dalian, China, in 2022. He is  pursuing the Ph.D. degree in Software Engineering at Dalian University of Technology, Dalian, China. His research interests include image processing and fusion.
	\end{IEEEbiography}
	\vspace{-1cm}
	\begin{IEEEbiography}[{\includegraphics[width=1in,height=1.25in,clip,keepaspectratio]{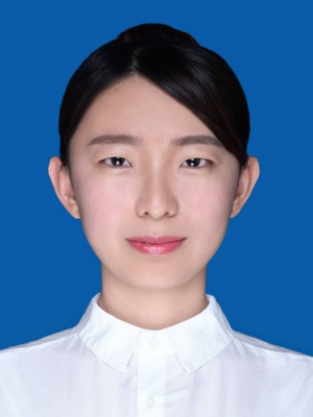}}] {Di Wang} received her M.S. degree in Computer Science and Engineering from Nanjing University of Science and Technology, Nanjing, China, in 2021. She is currently pursuing a Ph.D. degree in Software Engineering at Dalian University of Technology (DUT), Dalian, China. Her research interests include computer vision, image fusion, and deep learning.
	\end{IEEEbiography}
	\vspace{-1cm}
	\begin{IEEEbiography}[{\includegraphics[width=1in,height=1.25in,clip,keepaspectratio]{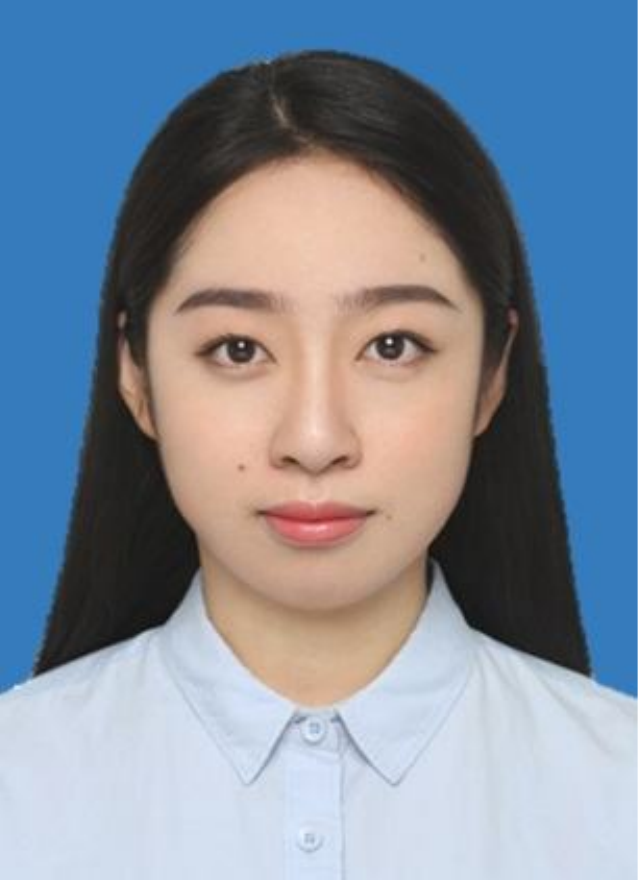}}] {Zhiying Jiang}received the B.E. degree in software engineering from Dalian Maritime University, China, in 2017, and the M.S. and Ph.D. degrees in software engineering from Dalian University of Technology, China, in 2020 and 2024, respectively. She is currently with the Key Laboratory for Ubiquitous Network and Service Software of Liaoning Province, Dalian University of Technology. Her research interests include computer vision, image restoration, and image stitching.
	\end{IEEEbiography}
	\vspace{-1cm}
	\begin{IEEEbiography}[{\includegraphics[width=1in,height=1.25in,clip,keepaspectratio]{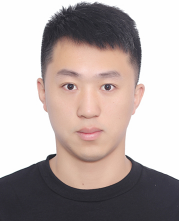}}] {Long Ma}received the Ph.D. degree in software engineering at Dalian University of Technology, Dalian, China, in 2019. He is currently a Postdoc the School of Software Technology, Dalian University of Technology (DUT), Dalian, China. His research interests include computer vision and deep learning.
	\end{IEEEbiography}
	\vspace{-1cm}
	\begin{IEEEbiography}[{\includegraphics[width=1in,height=1.25in,clip,keepaspectratio]{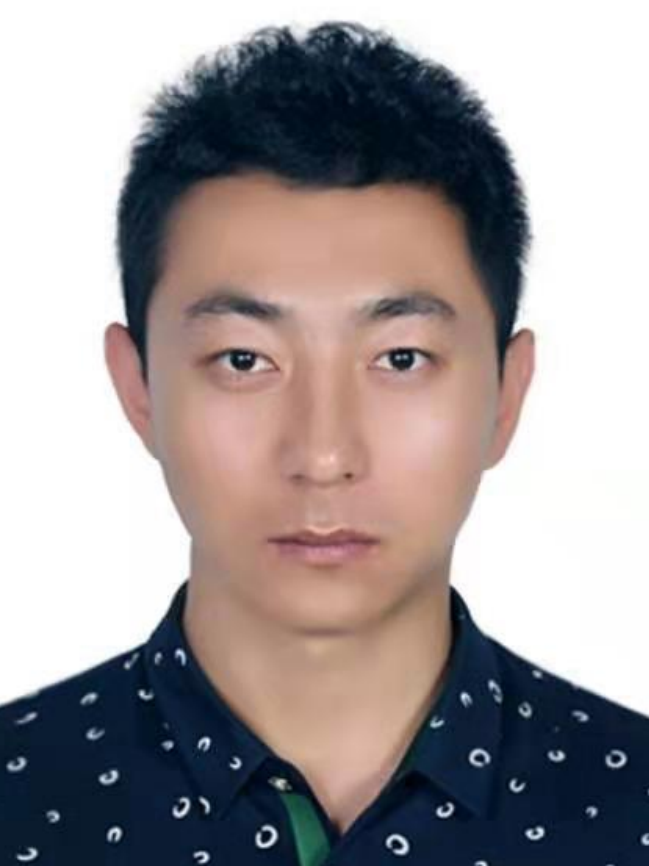}}] {Wei Zhong}received the B.S. degree from Dalian University of Technology in 2008 and his M.S. and Ph.D. degrees from Waseda University in 2010 and 2014, respectively. Currently, he is a Professor at Dalian University of Technology. His research focuses on computer vision, image processing, VLSI design, and hardware-software co-design.
	\end{IEEEbiography}
	\vspace{-1cm}
	\begin{IEEEbiography}[{\includegraphics[width=1in,height=1.25in,clip,keepaspectratio]{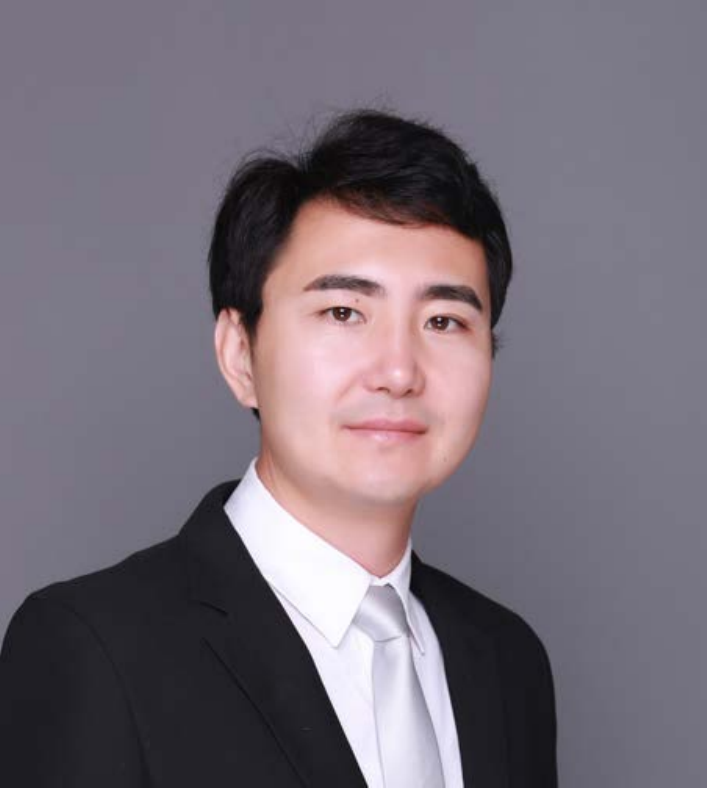}}] {Xin Fan} received the B.E. and Ph.D. degrees in information and communication engineering from Xian Jiaotong University, Xian, China, in 1998 and 2004, respectively. He was with Oklahoma State University, Stillwater, from 2006 to 2007, as a post-doctoral research fellow. He joined the School of Software, Dalian University of Technology, Dalian, China, in 2009. His current research interests include computational geometry and machine learning, and their applications to low-level image processing and DTI-MR image analysis.
	\end{IEEEbiography}
	\vspace{-1cm}
	\begin{IEEEbiography}[{\includegraphics[width=1in,height=1.25in,clip,keepaspectratio]{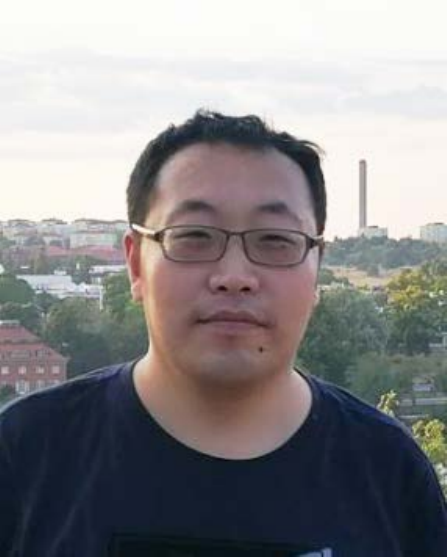}}] {Risheng Liu} received his B.S. (2007) and Ph.D. (2012) from Dalian University of Technology, China. He conducted joint Ph.D. research at Carnegie Mellon University from 2010 to 2012 and was a Hong Kong Scholar at The Hong Kong Polytechnic University from 2016 to 2018. He is currently a full professor in the DUT School of Software Technology. His research interests include optimization, computer vision, and multimedia.
	\end{IEEEbiography}
	\vspace{-1cm}

\end{document}